\pdfoutput=1
\documentclass[format=acmsmall, review=false, printccs=false, 
printacmref=false, printfolios=false, authorversion=false, 
timestamp=false, authordraft=false, anonymous=false, screen=false]{acmart}

\usepackage{amssymb}
\usepackage{color}
\usepackage{chngcntr}
\usepackage{algorithm}
\usepackage[noend]{algpseudocode}
\usepackage{listings}
\usepackage{soul}
\usepackage{amsmath}
\usepackage{mathtools}
\usepackage{multirow}
\usepackage{graphicx}
\usepackage[flushleft]{threeparttable}
\usepackage{booktabs}
\usepackage{flushend}
\usepackage{bm}
\usepackage{amsfonts}
\usepackage[T1]{fontenc}
\usepackage{enumitem}
\usepackage{tabularx}
\usepackage{array}
\usepackage{placeins}
\usepackage{rotating}
\usepackage{subcaption}
\usepackage{nameref}

\usepackage{float}
\floatstyle{plain}
\newfloat{lstfloat}{tbp}{lop}
\floatname{lstfloat}{Listing}
\usepackage{tikz}
\usepackage{tikz-qtree}
\usepackage[edges]{forest}
\usetikzlibrary{positioning, shapes, arrows.meta,shadows}

\usepackage{amsthm}
\usepackage{thmtools}
\declaretheorem[style=definition,qed={\footnotesize$\blacksquare$}]{definition}
\declaretheorem[style=definition]{proposition}

\usepackage{flushend}

\usepackage[mathcal,mathscr]{euscript}
\usepackage[frenchstyle,expert]{mathdesign}
\usepackage{eulervm}

\DeclareFontFamily{OT1}{pzc}{}
\DeclareFontShape{OT1}{pzc}{m}{it}{<-> s * [1.2] pzcmi7t}{}
\DeclareMathAlphabet{\mathpzc}{OT1}{pzc}{m}{it}

\newcommand{\mc}[1]{\mathcal{#1}}

\newcommand{\mz}[1]{\mathpzc{#1}}

\definecolor{red}{rgb}{1.00,0.00,0.00}
\definecolor{blue}{rgb}{0.00,0.00,1.00}
\definecolor{green}{rgb}{0.4,1.00,0.0}
\definecolor{yellow}{rgb}{0.5,0.5,0.0}
\definecolor{gray}{rgb}{0.5,0.5,0.5}
\definecolor{ipython_frame}{RGB}{207, 207, 207}
\definecolor{halfgray}{gray}{0.3}

\algrenewcommand\alglinenumber[1]{\tiny\color{black} #1~}

\lstdefinestyle{customlst}{
    mathescape,
    numbersep=-5.5pt,
    numberstyle=\tiny\color{halfgray},
    captionpos=b,
    stepnumber=1,
    basicstyle=\ttfamily\fontsize{7.5}{0}\selectfont,
    keywordstyle=\bfseries}
\makeatletter
\def\moverlay{\mathpalette\mov@rlay}
\def\mov@rlay#1#2{\leavevmode\vtop{%
   \baselineskip\z@skip \lineskiplimit-\maxdimen
   \ialign{\hfil$\m@th#1##$\hfil\cr#2\crcr}}}
\newcommand{\charfusion}[3][\mathord]{
    #1{\ifx#1\mathop\vphantom{#2}\fi
        \mathpalette\mov@rlay{#2\cr#3}
      }
    \ifx#1\mathop\expandafter\displaylimits\fi}
\makeatother

\newcommand{\figref}[1]{Figure~\ref{fig:#1}}

\newcommand{\lstref}[1]{Listing~\ref{lst:#1}}

\newcommand{\Half}{\frac{1}{2}}

\newcommand{\Init}{\texttt{init}}
\newcommand{\Static}{\texttt{static}}
\newcommand{\End}{\texttt{end}}
\newcommand{\Struc}{\texttt{Struct}}

\newcommand{\kjoin}{\sqcup}
\newcommand{\KJoin}{\bigsqcup}
\usepackage{xspace}
\newcommand{\ttf}[1]{{\texttt{#1}}}
\newcommand{\ttfs}[1]{{\small\texttt{#1}}}

\newcommand{\EBPD}{\Delta=(\mc{D}_a,\mc{D}_c,\mc{R},\mc{E},\mc{M})}
\newcommand{\Dom}{\mc{D}=(\mc{L},\Sigma,\mc{S},\mc{O})}
\newcommand{\ConDom}{\mc{D}_c=(\mc{L}_c,\Sigma_c,\mc{S}_c,\mc{O}_c)}
\newcommand{\AbsDom}{\mc{D}_a=(\mc{L}_a,\Sigma_a,\mc{S}_a,\mc{O}_a)}

\newcommand{\Scope}{\mz{S}}
\newcommand{\Sch}{m=(t,\Scope,\Omega)}

\newcommand{\Len}[1]{{\texttt{len$(#1)$}}}

\newcommand{\Nil}{\varnothing}

\newcommand{\StackDom}{\textsc{stacking-blocks}\xspace}
\newcommand{\Rover}{\textsc{rover}\xspace}
\newcommand{\Cafe}{\textsc{cafe}\xspace}

\begin{document}
\title[Experience-Based Planning Domains]
{Learning Task Knowledge and its Scope of Applicability\\in Experience-Based Planning Domains}

\author{Vahid Mokhtari}
\author{Lu\'{i}s Seabra Lopes}
\author{Armando J. Pinho}
\affiliation{%
  \institution{The University of Aveiro}
  \country{Portugal}}
\email{mokhtari.vahid@ua.pt}

\author{Roman Manevich}
\affiliation{%
  \institution{The University of Texas at Austin}
  \country{USA}
}
\email{romanm@cs.bgu.ac.il}

\begin{abstract}
\section*{ABSTRACT}
\emph{Experience-based planning domains}
(EBPDs) have been recently proposed to improve problem solving by learning
from experience.
EBPDs provide important concepts for long-term learning and planning in robotics.
They rely on acquiring and using task knowledge, i.e., activity schemata,
for generating concrete solutions to problem instances in a class of tasks.
Using Three-Valued Logic Analysis (TVLA), we extend previous work to generate
a set of conditions as the scope of applicability for an activity schema.
% Using Three-Valued Logic Analysis (TVLA), we extend previous work to generate
% a set of conditions, which is called the scope of applicability, for activity
% schemata.
The inferred scope is a bounded representation of a set of problems
of potentially unbounded size, in the form of a 3-valued logical structure,
which allows an EBPD system to automatically find an applicable activity
schema for solving task problems.
% We demonstrate the utility of our approach in a set of classes of problems
% in a simulated domain and real world tasks.
We demonstrate the utility of our approach in a set of classes of problems
in a simulated domain and a class of real world tasks in a fully physically
simulated PR2 robot in Gazebo.

\end{abstract}

\keywords{%Experience-based learning and planning,
Experience-based planning domains,
Learning from experience,
Robot task learning and planning,
%Conceptualization; Generalization; Abstraction; Loop detection;
Inferring scope of applicability.
%Robot task learning with loops,
%Robot task learning and planning,
%Robot tasks with loops
%Conceptualization
%Learning from demonstration
}

\maketitle

\renewcommand{\shortauthors}{V. Mokhtari et al.}

%%%%%%%%%%%%%%%%%%%%%%%%%%%%%%%%%%%%%%%%%%%%%%%%%%%%%%%%%%%
%%%%%%%%%%%%%%%%%%%%%%%%%%%%%%%%%%%%%%%%%%%%%%%%%%%%%%%%%%%
%%%%%%%%%%%%%%%%%%%%%%%%%%%%%%%%%%%%%%%%%%%%%%%%%%%%%%%%%%%
%%%%%%%%%%%%%%%%%%%%%%%%%%%%%%%%%%%%%%%%%%%%%%%%%%%%%%%%%%%
%%%%%%%%%%%%%%%%%%%%%%%%%%%%%%%%%%%%%%%%%%%%%%%%%%%%%%%%%%%
%%%%%%%%%%%%%%%%%%%%%%%%%%%%%%%%%%%%%%%%%%%%%%%%%%%%%%%%%%%
%%%%%%%%%%%%%%%%%%%%%%%%%%%%%%%%%%%%%%%%%%%%%%%%%%%%%%%%%%%
\section{INTRODUCTION}\label{sec:introduction}

Planning is a key ability for intelligent robots, increasing their autonomy 
and flexibility through the construction of sequences of actions to achieve 
their goals \cite{ghallab2004automated}.
Planning is a hard problem and even what is known historically as 
classical planning is PSPACE-complete over propositional state variables 
\cite{bylander1994computational}.
To carry out increasingly complex tasks, robotic communities make strong 
efforts on developing robust and sophisticated high-level decision making 
models and implement them as planning systems. 
One of the most challenging issues is to find an optimum in a trade-off 
between computational efficiency and needed domain expert engineering 
work to build a reasoning system. 
In a recent work \cite{mokhtari2016jint,mokhtari2016icaps,vahid2017prletter,vahid2017iros}, 
we have proposed and integrated the notion of \emph{Experience-Based 
Planning Domain} (EBPD)---a framework that integrates important concepts 
for long-term learning and planning---into robotics. 
An EBPD is an extension of the standard \emph{planning domains} which 
in addition to planning operators, includes experiences and methods 
(called \emph{activity schemata}) for solving classes of problems. 
Figure~\ref{fig:ebpd} illustrates the experience extraction, learning and 
planning pipeline for building an EBPD system. 
\emph{Experience extraction} provides a human-robot interaction for teaching 
tasks and an approach to recording experiences of past robot's observations 
and activities.
Experiences are used to learn activity schemata, i.e., methods of guiding a 
search-based planner for finding solutions to other related problems. 
\emph{Conceptualization} combines several techniques, including deductive 
generalization, different forms of abstraction, feature extraction, loop 
detection and inferring the scope of applicability, to generate activity 
schemata from experiences. \emph{Planning} is a hierarchical problem solver 
consisting of an abstract and a concrete planner which applies learned 
activity schemata for problem solving. 
In previous work, algorithms have been developed for experience extraction 
\cite{vahid2014experience,mokhtari2016jint}, activity schema learning and 
planning \cite{mokhtari2016jint,vahid2017prletter,vahid2017iros}.

%%%%%%%%%%%%%%%%%%%%%%%%%%%%%%%%%%%%%%%%%%%%%%%%%%%%%%%%%%%%%%%%%%%%%%%%%%%%%%%
\tikzstyle{box}=[rectangle, draw=black, 
               rectangle split, rectangle split parts=2,
               rounded corners, minimum width=5.5cm, minimum height=5.6cm]

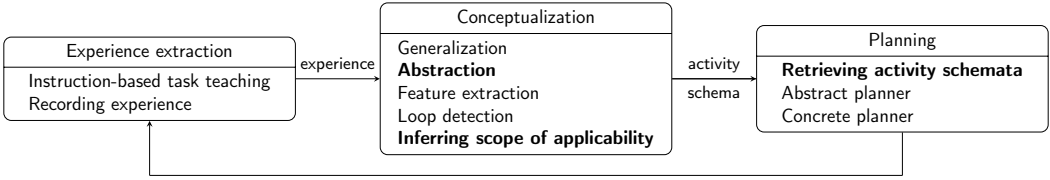
\begin{figure}[t]
\fontfamily{cmss}\selectfont
\begin{center}
\resizebox{\textwidth}{!}{%
\begin{tikzpicture}[node distance=1.6cm]

  \node[align=center] (teacher) [box] {
      {Experience extraction}
      \nodepart[align=left]{second}
      Instruction-based task teaching\\
      Recording experience
    };

  \node[align=center,right of=teacher, xshift=5.5cm] (learning) [box] {
      {Conceptualization}
      \nodepart[align=left]{second}
      Generalization\\
      \textbf{Abstraction}\\
      Feature extraction\\
      Loop detection\\
      \textbf{Inferring scope of applicability}
    };

  \node[align=center,right of=learning, xshift=5.5cm] (world) [box] {
      {Planning}
      \nodepart[align=left]{second}
      \textbf{Retrieving activity schemata}\\
      Abstract planner\\
      Concrete planner
    };

\node (fix1)    [below of=teacher, yshift=-.2cm] {};
\node (fix2)    [below of=world, yshift=-.2cm]{};

\path[->,>=stealth] 
         (teacher)     edge node[sloped,above] {\small experience} (learning)
         (learning)    edge node[sloped,above] {\small activity} (world);
\path[]  (learning)    edge node[sloped,below] {\small schema} (world);
\path[]  (world.south) edge [] (fix2.center)
         (fix2.center) edge [] (fix1.center);
\path[->,>=stealth] 
         (fix1.center) edge [->,>=stealth] (teacher.south);

\end{tikzpicture}}
\caption{An abstract illustration of the experience extraction, learning 
and planning pipeline in EBPDs.}
\label{fig:ebpd}
\end{center}
\end{figure}
%%%%%%%%%%%%%%%%%%%%%%%%%%%%%%%%%%%%%%%%%%%%%%%%%%%%%%%%%%%%%%%%%%%%%%%%%%%%%%%

In this paper, we present several recent improvements and extensions of 
EBPDs. The procedures highlighted in bold in Figure~\ref{fig:ebpd} outline 
the contribution of this paper. 
As the main contribution of this paper, we extend and improve the EBPDs 
framework to automatically retrieve an applicable activity schema 
for solving a task problem. %, among several learned activity schemata.
We propose an approach to infer a set of conditions from an experience 
that determines the \emph{scope of applicability} of an activity schema 
for solving a set of task problems. The inferred scope is a $3$-valued 
logical structure \cite{kleene1952introduction} (i.e., a structure that 
extends Boolean logic by introducing an indefinite value $\Half$ to denote 
either $0$ or $1$) which associates a bounded representation for a set of 
problems in the form of $2$-valued logical structures of potentially 
unbounded size. 
We employ Three-Valued Logic Analysis (TVLA) \cite{SRW:TOPLAS02} both 
to infer the scope of applicability of activity schemata 
(Section~\ref{sec:tvla_learning}) and to test whether existing activity 
schemata can be used to solve given task problems 
(Section~\ref{sec:tvla_execution}). 

We also extend and improve the abstraction methodology used in EBPDs 
(Section~\ref{sub:abstraction}). We propose to apply two independent 
abstraction hierarchies for reducing the level of detail during both 
learning an activity schema and planning, which leads to generate an 
abstract solution useful to reduce the search at the more concrete 
planning level. 

In the rest of the paper, we recapitulate the previous work and present 
an integrated and up-to-dated formal model of EBPDs 
(Section~\ref{sec:formalization}), and the approaches to learning 
activity schemata from robot experiences and task planning 
(Sections~\ref{sec:schema}--\ref{sec:planner}) 
(note that Sections~\ref{sec:schema} and \ref{sec:planner} are prior works).
A special attention is given to abstracting an experience and inferring the 
scope of applicability of an activity schema using the TVLA 
(Sections~\ref{sec:tvla_learning} and \ref{sec:tvla_execution}). 
We validate our system over a set of classes of problems in a simulated 
domain and a class of real world task in a fully physically simulated PR2 
robot in Gazebo (Section~\ref{sec:experiments}).

%%%%%%%%%%%%%%%%%%%%%%%%%%%%%%%%%%%%%%%%%%%%%%%%%%%%%%%%%%%
%%%%%%%%%%%%%%%%%%%%%%%%%%%%%%%%%%%%%%%%%%%%%%%%%%%%%%%%%%%
%%%%%%%%%%%%%%%%%%%%%%%%%%%%%%%%%%%%%%%%%%%%%%%%%%%%%%%%%%%
%%%%%%%%%%%%%%%%%%%%%%%%%%%%%%%%%%%%%%%%%%%%%%%%%%%%%%%%%%%
%%%%%%%%%%%%%%%%%%%%%%%%%%%%%%%%%%%%%%%%%%%%%%%%%%%%%%%%%%%
%%%%%%%%%%%%%%%%%%%%%%%%%%%%%%%%%%%%%%%%%%%%%%%%%%%%%%%%%%%
%%%%%%%%%%%%%%%%%%%%%%%%%%%%%%%%%%%%%%%%%%%%%%%%%%%%%%%%%%%

\section{RELATED WORK}\label{sec:literature}

The EBPDs' objective is to perform tasks.
Learning of Hierarchical Task Networks (HTNs) is among the most
related works to EBPDs.
In HTN planning, a plan is generated by decomposing a method for a given
task into simpler tasks until primitive tasks are reached that can be
directly achieved by planning operators.
CaMeL \cite{ilghami2002camel,ilghami2005learning} is an HTN learner which 
receives as input plan traces and the structure of an HTN method and tries 
to identify under which conditions the HTN is applicable. 
CaMeL requires all information about methods except for the preconditions.
The same group transcends this limitation in a later work \cite{ilghami2006hdl}
and presents the HDL algorithm which starts with no
prior information about the methods but requires hierarchical plan
traces produced by an expert problem-solver.
HTN-Maker \cite{hogg2008htn} generates an HTN domain model from a STRIPS 
domain model, a set of STRIPS plans, and a set of annotated tasks.
HTN-Maker generates and traverses a list of states by applying the actions
in a plan, and looks for an annotated task whose effects and preconditions
match some states. Then it regresses the effects of the annotated task
through a previously learned method or a new primitive task.
Overall, identifying a good hierarchical structure is an issue,
and most of the techniques in HTN learning rely on the hierarchical
structure of the HTN methods specified by a human expert.
On the contrary, the EBPDs framework presents a fully autonomous
approach to learning activity schemata with loops from single experiences.
The inclusion of loops in activity schemata is an alternative to recursive HTN methods.

Aranda~\cite{Srivastava2011615} takes a planning problem and finds 
a plan that includes loops. 
Using TVLA~\cite{LAmiS:SAS00} and back-propagation, Aranda finds 
an abstract state space from a set of concrete states of problem 
instances with varying numbers of objects
that guarantees completeness, i.e., the plan works for all
inputs that map onto the abstract state.
These strong guarantees come at a cost: (i) restrictions on the language
of actions; and (ii) high running times. Indeed computing the abstract state 
is worst-case doubly-exponential in the number of predicates.
In contrast, the EBPDs system assumes standard PDDL actions. We also use 
TVLA to compute an abstract structure that determines 
the scope of applicability of an activity 
schema, however, we trade completeness for a polynomial time algorithm, 
which results in dramatically better performance.

Loop\textsc{Distill} \cite{Winner07loopdistill} also 
learns plans with loops from example plans.
It identifies the largest matching sub-plan in a given example
and converts the repeating occurrences of the sub-plans into
a loop. The result is a domain-specific planning program
(dsPlanner), i.e., a plan with if-statements and while-loops
that can solve similar problems of the same class.
Loop\textsc{Distill}, nonetheless, does not address the applicability test of plans.

Other approaches in AI planning including
case based planning \cite{hammond1986chef,borrajo2015acm},
and macro operators \cite{fikes1972strips2,chrpa2010generation}
are also related to our work.
These methods tend to suffer from the utility problem, in which
learning more information can be counterproductive due to
the difficulty with storage and management of the information
and with determining which information should be used to solve a
particular problem.
In EBPDs, by combining generalization with abstraction in task learning,
it is possible to avoid saving large sets of concrete cases.
Additionally, since in EBPDs, task learning is supervised,
solving the utility problem can be to some extent delegated to
the user, who chooses which tasks and associated procedures to teach.

Other related work includes Learning from Demonstration (LfD) which puts 
effort into learning robot control programs by simply showing robots 
how to achieve tasks \cite{Argall2009469,billard2008robot}. 
This has the immediate advantage of requiring no specialized skill or training, 
and makes use of a human demonstrator's knowledge to identify which control 
program to acquire, typically by regression-based methods.
Although LfD is useful in learning primitive action-control policies (such 
as for object manipulation), it is unsuitable for learning complex tasks. 
LfD usually requires many examples in order to induce the intended control 
structure \cite{allen2007plow}. 
Moreover, the representations are task-specific and are not 
likely to transfer to structurally similar tasks \cite{chao2011towards}.

%%%%%%%%%%%%%%%%%%%%%%%%%%%%%%%%%%%%%%%%%%%%%%%%%%%%%%%%%%%
%%%%%%%%%%%%%%%%%%%%%%%%%%%%%%%%%%%%%%%%%%%%%%%%%%%%%%%%%%%
%%%%%%%%%%%%%%%%%%%%%%%%%%%%%%%%%%%%%%%%%%%%%%%%%%%%%%%%%%%
%%%%%%%%%%%%%%%%%%%%%%%%%%%%%%%%%%%%%%%%%%%%%%%%%%%%%%%%%%%
%%%%%%%%%%%%%%%%%%%%%%%%%%%%%%%%%%%%%%%%%%%%%%%%%%%%%%%%%%%
%%%%%%%%%%%%%%%%%%%%%%%%%%%%%%%%%%%%%%%%%%%%%%%%%%%%%%%%%%%
%%%%%%%%%%%%%%%%%%%%%%%%%%%%%%%%%%%%%%%%%%%%%%%%%%%%%%%%%%%

%%%%%%%%%%%%%%%%%%%%%%%%%%%%%%%%%%%%%%%%%%%%%%%%%%%%%%%%%%%
\section{RUNNING EXAMPLE}
\label{sec:running_example}

We develop a \StackDom planning domain, based on the blocks 
world domain, for representing provided concepts and definitions. 
Assume a set of blocks of red and blue colors sitting on a table. 
The goal is to build a vertical stack of red and blue blocks. 
The state of a problem in this domain consists of predicates with 
the following meanings. 
\texttt{pile(x)},
\texttt{table(x)},
\texttt{red(x)},
\texttt{blue(x)},
\texttt{pallet(x)}: \texttt{x} is a pile, table, red block, 
blue block, or pallet, respectively.
\texttt{attached(p,l)}: pile \texttt{p} is attached to location \texttt{l}.
\texttt{belong(h,l)}: hoist \texttt{h} belongs to location \texttt{l}.
\texttt{at(h,p)}: hoist \texttt{h} is at place \texttt{p}.
\texttt{holding(h,x)}: hoist \texttt{h} is holding block \texttt{x}.
\texttt{empty(h)}: hoist \texttt{h} is empty.
\texttt{on(x,y)}: block \texttt{x} is on block \texttt{y}.
\texttt{ontable(x,t)}: block \texttt{x} is on table \texttt{t}.
\texttt{top(x,p)}: block \texttt{x} is the top of pile \texttt{p}.

The \StackDom domain has the actions with the following meanings. 
\texttt{move(h,x,y,l)}: 
hoist \texttt{h} moves from place \texttt{x} to place \texttt{y} 
at location \texttt{l}. 
\texttt{unstack(h,x,y,p,l)}: hoist \texttt{h} unstacks block \texttt{x} 
from block \texttt{y} on pile \texttt{p} at location \texttt{l}. 
\texttt{stack(h,x,y,p,l)}: hoist \texttt{h} puts block \texttt{x} 
on block \texttt{y} on pile \texttt{p} at location \texttt{l}. 
\texttt{pickup(h,x,t,l)}: hoist \texttt{h} picks up block \texttt{x} 
from table \texttt{t} at location \texttt{l}. 
\texttt{putdown(h,x,t,l)}: hoist \texttt{h} puts down block \texttt{x} 
on table \texttt{t} at location \texttt{l}.

We define a specific class of `stack' problems where an equal number of 
red and blue blocks are initially on a table and need to be stacked on a 
pile with blue blocks at bottom and red blocks on top using a hoist 
which can hold only one block at a time. 
Generalizing from this example, we formally present and define concepts 
used for creating this domain and problem solving in EBPDs. 

%%%%%%%%%%%%%%%%%%%%%%%%%%%%%%%%%%%%%%%%%%%%%%%%%%%%%%%%%%%
\section{A FORMAL MODEL OF EXPERIENCE-BASED PLANNING DOMAINS}
\label{sec:formalization}

An EBPD is a unified framework that provides intelligent robots with 
the capability of problem solving by learning from experience 
\cite{mokhtari2016jint,vahid2017prletter}. Problem solving in this 
framework is achieved using a hierarchical problem solver, consisting 
of an abstract and a concrete planning domain, which employs a set of 
learned activity schemata for guiding a search-based planning. 

%%%%%%%
Formally, an EBPD is described as a tuple 
$\EBPD$ 
where 
$\mc{D}_a$ is an abstract planning domain, 
$\mc{D}_c$ is a concrete planning domain, 
$\mc{R}$ is a set of abstraction hierarchies (i.e., inference rules) 
$f:\mc{D}_c\to\mc{D}_a$ to translate the concrete space in $\mc{D}_c$ 
into the abstract space in $\mc{D}_a$, 
$\mc E$ is a set of experiences, 
and $\mc M$ is a set of methods in the form of activity schemata for 
solving problems. 
%%%%%%%

In general, a planning domain of problem solving $\Dom$ is described by 
a first-order logic language $\mc{L}$, a finite subset of ground atoms 
$\Sigma$ of $\mc{L}$ for representing the static or invariant properties 
of the world (properties of the world that are always true), a set of all 
possible states $\mc{S}$, in which every state $s\in\mc{S}$ is a set of 
ground atoms of $\mc{L}$ representing dynamic or transient properties of 
the world (i.e., $s\cap\Sigma\neq\emptyset$), and a set of planning 
operators $\mc O$. { In EBPDs, the abstract and concrete planning 
domains are denoted by $\AbsDom$ and $\ConDom$ respectively.}

%%%%%%%%%%%%%%%%%%%%%%%%%%%%%%%%%%%%%%%%%%%%%%%%%%%%%%%%%%%
A \emph{planning operator} $o \in\mc{O}$ is described as a tuple 
$( h,S,P,E )$ where 
$h$ is the planning operator head, 
$S$ is the static precondition of $o$, a set of predicates that must 
be proved in $\Sigma$,  
$P$ is the precondition of $o$, a set of literals that must be proved 
in a state $s \in \mc{S}$ in order to apply $o$ in $s$, and 
$E$ is the effect of $o$, a set of literals specifying the changes on $s$ 
effected by $o$.
A head takes a form \(n(x_1,...,x_k)\) in which 
$n$ is the name and \(x_1,...,x_k\) are the arguments, e.g., 
{\texttt{(pick ?block ?table)}} 
\footnote{ The notation in the Planning Domain Definition Language 
(PDDL) is used to represent EBPDs. All terms starting with a question 
mark (?) are variables, and the rest are constants or function symbols.}.
Any ground instance of a planning operator is called an \emph{action}.
\lstref{absoperator} shows a planning operator in EBPDs.
%%%%%%%%%%%%%%%%%%%%%%%%%%%%%%%%%%%%%%%%%%%%%%%%%%%%%%%%%%%

%%%%%%%%%%%%%%%%%%%%%%%%%%%%%%%%%%%%%%%%%%%%%%%%%%%%%%%%%%%
\begin{lstlisting}[style=customlst,
    label=lst:absoperator,
    float,
    % belowskip=-8pt,
    morekeywords={action, static, parameters, precondition, effect},
    caption={Representation of a planning operator in EBPDs.}]
(:action pick
 :parameters   (?block ?table)
 :static       (and (table ?table)(block ?block))
 :precondition (ontable ?block ?table)
 :effect       (and (holding ?block)(not(ontable ?block ?table))))
\end{lstlisting}
%%%%%%%%%%%%%%%%%%%%%%%%%%%%%%%%%%%%%%%%%%%%%%%%%%%%%%%%%%%

Abstraction in EBPDs is achieved by dropping or transforming predicates 
and operators of the concrete planning domain $\mc{D}_c$ into the abstract 
planning domain $\mc{D}_a$. This transformation involves two independent 
abstraction hierarchies: a \emph{predicate abstraction hierarchy} and an 
\emph{operator abstraction hierarchy} which are expressed in $\mc{R}$. 
The predicate abstraction hierarchy is a set of abstraction relations, 
each one relating a concrete predicate 
$p_c(u_1,\dots,\allowbreak u_n)\allowbreak \in \mc{L}_c$ to an abstract 
predicate $p_a(v_1,\dots,v_m) \in \mc{L}_a$, such that $m \leq n$ and 
$(v_1,\dots,\allowbreak v_m) \allowbreak \subseteq (u_1,\dots,u_n)$; 
or to $\Nil$ ($nil$).
That is, a concrete predicate in $\mc{L}_c$ might: map onto 
an abstract predicate in $\mc{L}_a$ by replacing predicate symbols and 
excluding some arguments of the concrete predicate from the arguments of 
the abstract predicate, e.g., 
{\texttt{(holding ?hoist ?block) $\to$ (holding ?block)}}; 
or map onto $\varnothing$ ($nil$), that is, it is excluded 
from the abstract predicates, e.g., 
{\texttt{(attached ?pile ?location) $\to \varnothing$}}. 
Similarly, the operator abstraction hierarchy translates concrete 
operators in $\mc{O}_c$ into abstract operators in $\mc{O}_a$. 
In this abstraction, a concrete operator in $\mc{O}_c$ might: map onto 
an abstract operator in $\mc{O}_a$ by replacing operator symbols and 
excluding some arguments of the concrete operator from the arguments 
of the abstract operator, e.g., 
{\texttt{(pickup ?hoist ?block ?table ?loc) $\to$ (pick ?block ?table)}};
or map onto $\varnothing$ ($nil$), that is, it is excluded 
from the abstract operators, e.g., 
{\texttt{(move ?hoist ?from ?to ?loc) $\to \varnothing$}}. 

In this paper, a functional expression \texttt{parent}$(x)$, wherever it 
is used, given a concrete predicate/operator, returns the parent of $x$, 
i.e., an abstract predicate/operator corresponding to the concrete 
predicate/operator.
Tables~\ref{tbl:predicate_hierarchies} and \ref{tbl:operator_hierarchies} 
present the predicate and operator abstraction hierarchies in the 
\StackDom EBPD. 
\footnote{ As a prerequisite of EBPDs, it is assumed that descriptions 
of the abstract and concrete planning domains $(\mc{D}_a,\mc{D}_c)$ with 
operators and predicates abstraction hierarchies $\mc{R}$ are given by a 
domain expert. Although it may require more effort to specify the abstract 
language, but we believe this is a price we have to pay to make planning 
more tractable in certain situation. Moreover, automatic definition of 
abstract and concrete planning domains is beyond the scope of this work.}

%%%%%%%%%%%%%%%%%%%%%%%%%%%%%
\begin{table}
\centering
\caption{Predicate abstraction hierarchy in the \StackDom EBPD.}
\label{tbl:predicate_hierarchies}
% \vspace{5pt}
\setlength\extrarowheight{-1pt}
{\footnotesize
\begin{tabular}{rl}
% \hline
\textbf{\normalsize Abstract predicate} & \textbf{\normalsize Concrete predicate}
\smallskip\\
\hline
\texttt{(table ?table)}          & \texttt{(table ?table)} \\
\texttt{(pile ?pile)}            & \texttt{(pile ?pile)} \\
\texttt{(block ?block)}          & \texttt{(block ?block)} \\
\texttt{(blue ?block)}           & \texttt{(blue ?block)} \\
\texttt{(red ?block)}            & \texttt{(red ?block)} \\
\texttt{(pallet ?pallet)}        & \texttt{(pallet ?pallet)} \\
\texttt{(on ?block1 ?block2)}    & \texttt{(on ?block1 ?block2)} \\
\texttt{(ontable ?block ?table)} & \texttt{(ontable ?block ?table)} \\
\texttt{(top ?block ?pile)}      & \texttt{(top ?block ?pile)} \\
\texttt{(holding ?block)}        & \texttt{(holding ?hoist ?block)} \\
$\varnothing$\;                  & \texttt{(location ?location)} \\
$\varnothing$\;                  & \texttt{(hoist ?hoist)} \\
$\varnothing$\;                  & \texttt{(attached ?pile ?location)} \\
$\varnothing$\;                  & \texttt{(belong ?hoist ?location)} \\
$\varnothing$\;                  & \texttt{(at ?hoist ?pile)} \\
$\varnothing$\;                  & \texttt{(empty ?hoist)} \\
% \hline
\end{tabular}}
\end{table}
%%%%%%%%%%%%%%%%%%%%%%%%%%%%%

%%%%%%%%%%%%%%%%%%%%%%%%%%%%%
\begin{table}[t]
% \setlength{\tabcolsep}{12pt}
%\footnotesize
\centering
\caption{Operator abstraction hierarchy in the \StackDom EBPD.}
\label{tbl:operator_hierarchies}
% \setlength\extrarowheight{-1pt}
% \vspace{5pt}
% \resizebox{\columnwidth}{!}
{\footnotesize
\begin{tabular}{@{}rl@{}}
% \hline
\textbf{\normalsize Abstract operator} & \textbf{\normalsize Concrete operator}
\smallskip\\
\hline
\texttt{(unstack ?block1 ?block2 ?pile)} & \texttt{(unstack ?hoist ?block1 ?block2 ?pile ?loc)} \\
\texttt{(stack ?block2 ?block1 ?pile)}   & \texttt{(stack ?hoist ?block2 ?block1 ?pile ?loc)} \\
\texttt{(pick ?block ?table)}            & \texttt{(pickup ?hoist ?block ?table ?loc)} \\
\texttt{(put ?block ?table)}             & \texttt{(putdown ?hoist ?block ?table ?loc)} \\
$\varnothing$\;                          & \texttt{(move ?hoist ?from ?to ?loc)} \\
% \hline
\end{tabular}}
\end{table}
%%%%%%%%%%%%%%%%%%%%%%%%%%%%%

We propose to use experience given in the form of a concrete previously 
solved problem and to abstract this experience for its reuse in new 
situations. 
%%%%%%%
An \emph{experience} $e \in\mc{E}$ is a triple of ground structures 
$( t,K,\pi )$ where $t$ is a task achieved in the experience, i.e., a 
functional expression of the form $n(c_1,...,c_k)$ with $n$ being the 
task name and each $c_i$ a constant, e.g., {\ttf{(stack table1 pile1)}}, 
$K$ is a set of key-properties describing properties of the world in 
the experience, and $\pi$ is a solution plan to achieve $t$. 
Every \emph{key-property} is of the form $\tau(p)$ where $\tau$ is a 
temporal symbol and $p$ is a predicate. Temporal symbols specify the 
temporal extent of predicates in the experience. Three types of temporal 
symbols are used in key-properties, namely 
\texttt{init}---true at the initial state, 
\texttt{static}---always true during an experience, and
\texttt{end}---true at the final state, 
e.g., {\ttf{(end(top block8 pile1))}}. 
\lstref{experience} shows part of an experience in EBPDs.
%%%%%%%

Experiences are collected through human-robot interaction and 
instruction-based teaching. We previously presented methods and 
approaches of instructing and teaching a robot how to achieve a 
task as well as extracting and recording experiences 
\cite{vahid2014experience,mokhtari2016jint}. Experience extraction 
is beyond the scope of this paper. 

%%%%%%%%%%%%%%%%%%%%%%%%%%%%%
\begin{figure}[!t]
\begin{lstlisting}[style=customlst,
    % float=t,
    label=lst:experience,
    % abovecaptionskip=10pt,
    morekeywords={define, experience, domain, episode_id,
    task, parameters, key, properties, plan, objects},
    caption={Part of the `stack' experience in the \StackDom EBPD. 
    There are 8 (4 blue and 4 red) blocks in this experience. 
    The goal of the task in this experience is to stacking the blocks from 
    a table on a pile. 
    The key-properties describe the initial, final and static world information 
    of the experience (some key-properties are omitted due to limited space). 
    The plan solution to this problem contains 31 primitive actions. 
    }]
(:task stack
 :parameters     (table1 pile1)
 :key-properties ((static(pile pile1))
                  (static(table table1))
                  (static(location location1))
                  (static(hoist hoist1))
                  (static(attached pile1 location1))
                  (static(attached table1 location1))
                  (static(belong hoist1 location1))
                  (static(pallet pallet1))
                  (static(block block1))
                  (static(block block2))
                  (static(block block3))
                   $\cdots$
                  (static(block block6))
                  (static(block block7))
                  (static(block block8))
                  (static(blue block1))
                  (static(blue block2))
                  (static(blue block3))
                   $\cdots$
                  (static(red block6))
                  (static(red block7))
                  (static(red block8))
                  (init(top pallet1 pile1))
                  (init(ontable block1 table1))
                  (init(ontable block2 table1))
                  (init(ontable block3 table1))
                   $\cdots$
                  (init(ontable block6 table1))
                  (init(ontable block7 table1))
                  (init(ontable block8 table1))
                  (init(at hoist1 table1))
                  (init(empty hoist1))
                  (end(on block1 pallet1))
                  (end(on block2 block1))
                  (end(on block3 block2))
                   $\cdots$
                  (end(on block6 block5))
                  (end(on block7 block6))
                  (end(on block8 block7))
                  (end(top block8 pile1))
                  (end(at hoist1 pile1))
                  (end(empty hoist1)))
 :plan           ((pickup hoist1 block1 table1 location1)
                  (move hoist1 table1 pile1 location1)
                  (stack hoist1 block1 pallet1 pile1 location1)
                  (move hoist1 pile1 table1 location1)
                  (pickup hoist1 block2 table1 location1)
                  (move hoist1 table1 pile1 location1)
                  (stack hoist1 block2 block1 pile1 location1)
                  (move hoist1 pile1 table1 location1)
                  (pickup hoist1 block3 table1 location1)
                  (move hoist1 table1 pile1 location1)
                  (stack hoist1 block3 block2 pile1 location1)
                  (move hoist1 pile1 table1 location1)
                  (pickup hoist1 block4 table1 location1)
                  (move hoist1 table1 pile1 location1)
                  (stack hoist1 block4 block3 pile1 location1)
                  (move hoist1 pile1 table1 location1)
                  (pickup hoist1 block5 table1 location1)
                  (move hoist1 table1 pile1 location1)
                  (stack hoist1 block5 block4 pile1 location1)
                  (move hoist1 pile1 table1 location1)
                  (pickup hoist1 block6 table1 location1)
                  (move hoist1 table1 pile1 location1)
                  (stack hoist1 block6 block5 pile1 location1)
                  (move hoist1 pile1 table1 location1)
                  (pickup hoist1 block7 table1 location1)
                  (move hoist1 table1 pile1 location1)
                  (stack hoist1 block7 block6 pile1 location1)
                  (move hoist1 pile1 table1 location1)
                  (pickup hoist1 block8 table1 location1)
                  (move hoist1 table1 pile1 location1)
                  (stack hoist1 block8 block7 pile1 location1)))
\end{lstlisting}
\vspace{-30pt}
\end{figure}
%%%%%%%%%%%%%%%%%%%%%%%%%%%%%

Extracted experiences are the main inputs to acquire activity schemata.
Activity schemata are task planning knowledge obtained from experiences 
and contain generic solutions to classes of task problems. 
%%%%%%%
An \emph{activity schema} $m \in\mc{M}$ is a triple of ungrounded 
structures $\Sch$, where $t$ is the target task to perform by a robot, 
e.g., {\texttt{(stack ?table ?pile)}}, $\Scope$ is the scope of 
applicability of $m$, and $\Omega$ is a sequence of enriched abstract 
operators (also called en enriched abstract plan). Each \emph{enriched 
abstract operator}, denoted by $\omega$, is a pair $(o, F)$, where $o$ 
is an abstract operator head, and $F$ is a set of features of $o$, 
i.e., ungrounded key-properties, obtained from an experience, that 
characterize $o$.
%%%%%%%
In Section~\ref{sec:schema}, we present the method of learning and a 
concrete example of an activity schema. In Section~\ref{sec:tvla_learning}, 
we further develop the definition of the scope of applicability and present 
a method of inferring the scope of applicability for an activity schema 
from an experience.

%%%%%%%
Finally, a \emph{task planning problem} in EBPDs is described as a tuple 
of ground structures $\mc P=( t,\sigma,s_0,g )$
where $t$ is the target task, %e.g., \texttt{(clear table1)},
$\sigma\subseteq\Sigma$ is a subset of the static world information, 
$s_0$ is the initial world state, and $g$ is the goal. 
\footnote{
A full representation and implementation of the 
\StackDom EBPD, and a set of all concepts required for 
problem-solving in this EBPD are available at: \url{https://github.com/mokhtarivahid/ebpd/tree/master/domains/}.}
%%%%%%%

%%%%%%%%%%%%%%%%%%%%%%%%%%%%%%%%%%%%%%%%%%%%%%%%%%%%%%%%%%%
%%%%%%%%%%%%%%%%%%%%%%%%%%%%%%%%%%%%%%%%%%%%%%%%%%%%%%%%%%%
%%%%%%%%%%%%%%%%%%%%%%%%%%%%%%%%%%%%%%%%%%%%%%%%%%%%%%%%%%%
%%%%%%%%%%%%%%%%%%%%%%%%%%%%%%%%%%%%%%%%%%%%%%%%%%%%%%%%%%%
%%%%%%%%%%%%%%%%%%%%%%%%%%%%%%%%%%%%%%%%%%%%%%%%%%%%%%%%%%%
%%%%%%%%%%%%%%%%%%%%%%%%%%%%%%%%%%%%%%%%%%%%%%%%%%%%%%%%%%%
%%%%%%%%%%%%%%%%%%%%%%%%%%%%%%%%%%%%%%%%%%%%%%%%%%%%%%%%%%%

\section{LEARNING ACTIVITY SCHEMATA}\label{sec:schema}

In this section, we recapitulate the procedure for learning an 
activity schema in EBPDs. We also improve the abstraction methodology in 
EBPDs to achieve more compact and applicable concepts. 

%%%%%%%%%%%%%%%%%%%%%%%%%%%%%%%%%%%%%%%%%%%%%%%%%%%%%%%%%%%
% \smallskip
% \textbf{Generalization.} 
\subsection{Genralization} 
\label{sub:genralization}

The first stage, applied to an experience, in order to extract its basic 
principles, is a deductive generalization method based on the tradition 
of PLANEX \citep{fikes1972strips2} and Explanation-Based Generalization 
(EBG) \citep{mitchell1986explanation}. 
Through the generalization, a general concept is formulated from a single 
experience and domain knowledge. 
The proposed EBG method is carried out over the plan of the experience. 
In this transformation, constants appearing in the plan are replaced 
with variables, hence the plan becomes free from the specific constants 
and could be used in situations involving arbitrary constants. 
The EBG method consistently variablizes all constants appearing in the 
actions of the plan in the experience and when it gets the last action in 
the plan, propagates the variables for constants in the whole experience, 
i.e., the constants in the key-properties of the experience are also 
replaced with the variables obtained by the EBG. EBG then generates a 
generalized experience, i.e., a new planning control knowledge, which 
forms the basis of a learned activity schema.

%%%%%%%%%%%%%%%%%%%%%%%%%%%%%%%%%%%%%%%%%%%%%%%%%%%%%%%%%%%
\subsection{Abstraction} 
\label{sub:abstraction}

We propose to use an abstraction methodology for translating the obtained 
generalized experience into an abstracted generalized experience. Abstract 
representation allows, during problem solving, to solve given problems with 
a reduced computational effort. It also makes the learned concepts broader, 
more compact and widely applicable.
Given the predicate and operator abstraction hierarchies $\mc{R}$, the 
abstraction of an experience is achieved by transforming the concrete 
predicates and operators into abstract predicates and operators, which 
results in reducing the level of detail in the generalized experience. 
The predicate and operator abstraction hierarchies in $\mc{R}$ specify 
which of the concrete predicates/operators are mapped and which are skipped.

A concrete (generalized) experience $e=(t,K,\pi)$ is translated into an 
abstracted (generalized) experience 
$e_a=(t,K_a,\pi_a)$, denoted by $\ttf{Abs}(e)$, as follows:
\[
K_a   = \{\tau(\ttf{parent}(p)) \mid \tau(p) \in K\}, \quad
\pi_a = \{\ttf{parent}(o) \mid o \in \pi\}.
\]

Listing~\ref{lst:experience_gen} partially shows an experience after the 
generalization and abstraction. In this example, the abstraction is based 
on the predicate and operator abstraction hierarchies presented in 
Tables~\ref{tbl:predicate_hierarchies} and \ref{tbl:operator_hierarchies}.

%%%%%%%%%%%%%%%%%%%%%%%%%%%%%
\begin{lstlisting}[style=customlst,
    float=t,
    label=lst:experience_gen,
    belowskip=0pt,
    % abovecaptionskip=10pt,
    morekeywords={define, experience, domain, episode_id,
    task, parameters, key, properties, plan, objects},
    caption={After generalization and abstraction, the constants are replaced 
    with variables (Generalization), and some key-properties and actions are 
    excluded from the generalized experience (Abstraction) as specified in 
    the predicate and operator abstraction hierarchies.}]
(:task stack
 :parameters     (?table ?pile)
 :key-properties ((static(pile ?pile))
                  (static(table ?table))
                  (static(pallet ?pallet))
                  (static(block ?block1))
                  (static(block ?block2))
                   $\cdots$
                  (static(block ?block7))
                  (static(block ?block8))
                  (static(blue ?block1))
                  (static(blue ?block2))
                   $\cdots$
                  (static(red ?block7))
                  (static(red ?block8))
                  (init(top ?pallet ?pile))
                  (init(ontable ?block1 ?table))
                  (init(ontable ?block2 ?table))
                   $\cdots$
                  (init(ontable ?block7 ?table))
                  (init(ontable ?block8 ?table))
                  (end(on ?block1 ?pallet))
                  (end(on ?block2 ?block1))
                   $\cdots$
                  (end(on ?block7 ?block6))
                  (end(on ?block8 ?block7))
                  (end(top ?block8 ?pile)))
 :plan           ((!pick ?block1 ?table)
                  (!stack ?block1 ?pallet ?pile)
                  (!pick ?block2 ?table)
                  (!stack ?block2 ?block1 ?pile)
                  (!pick ?block3 ?table)
                  (!stack ?block3 ?block2 ?pile)
                  (!pick ?block4 ?table)
                  (!stack ?block4 ?block3 ?pile)
                  (!pick ?block5 ?table)
                  (!stack ?block5 ?block4 ?pile)
                  (!pick ?block6 ?table)
                  (!stack ?block6 ?block5 ?pile)
                  (!pick ?block7 ?table)
                  (!stack ?block7 ?block6 ?pile)
                  (!pick ?block8 ?table)
                  (!stack ?block8 ?block7 ?pile)))
\end{lstlisting}
%%%%%%%%%%%%%%%%%%%%%%%%%%%%%

%%%%%%%%%%%%%%%%%%%%%%%%%%%%%%%%%%%%%%%%%%%%%%%%%%%%%%%%%%%
% \smallskip
% \textbf{Feature extraction.}
\subsection{Feature extraction} 
\label{sub:features}

The discovery of meaningful features can contribute to the creation of 
a more concise and accurate learned concept \citep{fawcett1992automatic}. 
While abstraction reduces the level of detail in an experience, 
extracting other features would help to capture the essence of the 
experience. 
Features are properties of abstract operators in learned planning knowledge. 
In an experience, a \emph{feature} of an abstract operator is a 
key-property $\tau(p)$ such that $p$ contains at least one argument 
of the abstract operator and at least one argument of the task in the 
experience, that is, the feature links the abstract operator with the 
task in the experience. 
For example in Listing~\ref{lst:experience_gen}, the key-property 
\texttt{(init(ontable ?block1 ?table))} is a feature connecting 
\texttt{?block1}, the argument of an abstract operator \texttt{pick}, 
to \texttt{?table}, the argument of the task `stack'. 
For each abstract operator in a generalized and abstracted experience, 
all possible relations between the arguments of the abstract operator 
and the task arguments are automatically extracted and associated to 
the abstract operator. 
Features are intended to improve the performance of problem solving by 
guiding a planner toward a goal state and reducing the probability of 
backtracking, that is, during problem solving, objects that satisfy the 
features are preferable to instantiate actions. 
Listing~\ref{lst:schema} shows thus far the learned activity schema 
from the `stack' experience, after generalization, abstraction and 
feature extraction.

%%%%%%%%%%%%%%%%%%%%%%%%%%%%%
\begin{lstlisting}[style=customlst, 
    label=lst:schema, 
    float=t, 
    morekeywords={parameters, domain, define, method,
    activity, schema, abstract, plan, objects},
    caption={A learned activity schema thus far for the `stack' 
    task after generalization, abstraction and feature extraction. 
    Each abstract operator is associated with a set of features 
    (some are omitted due to limited space) that during problem 
    solving determine which objects can be used to instantiate 
    abstract actions. 
    }]
(:method stack
 :parameters (?table ?pile)
 :abstract-plan ((!pick ?block1 ?table)
                     ((static(blue ?block1))$\cdots$)
                 (!stack ?block1 ?pallet ?pile)
                     ((static(blue ?block1))(static(pallet ?pallet))$\cdots$)
                 (!pick ?block2 ?table)
                     ((static(blue ?block2))$\cdots$)
                 (!stack ?block2 ?block1 ?pile)
                     ((static(blue ?block1))(static(blue ?block2))$\cdots$)
                 (!pick ?block3 ?table)
                     ((static(blue ?block3))$\cdots$)
                 (!stack ?block3 ?block2 ?pile)
                     ((static(blue ?block2))(static(blue ?block3))$\cdots$)
                 (!pick ?block4 ?table)
                     ((static(blue ?block4))$\cdots$)
                 (!stack ?block4 ?block3 ?pile)
                     ((static(blue ?block3))(static(blue ?block4))$\cdots$)
                 (!pick ?block5 ?table)
                     ((static(red ?block5))$\cdots$)
                 (!stack ?block5 ?block4 ?pile)
                     ((static(blue ?block4))(static(red ?block5))$\cdots$)
                 (!pick ?block6 ?table)
                     ((static(red ?block6))$\cdots$)
                 (!stack ?block6 ?block5 ?pile)
                     ((static(red ?block5))(static(red ?block6))$\cdots$)
                 (!pick ?block7 ?table)
                     ((static(red ?block7))$\cdots$)
                 (!stack ?block7 ?block6 ?pile)
                     ((static(red ?block6))(static(red ?block7))$\cdots$)
                 (!pick ?block8 ?table)
                     ((static(red ?block8))(end(top ?block8 ?pile))$\cdots$)
                 (!stack ?block8 ?block7 ?pile)
                     ((static(red ?block7))(static(red ?block8))$\cdots$)))
\end{lstlisting}
%%%%%%%%%%%%%%%%%%%%%%%%%%%%%

%%%%%%%%%%%%%%%%%%%%%%%%%%%%%%%%%%%%%%%%%%%%%%%%%%%%%%%%%%%
% \smallskip
% \textbf{Loop detection.}
\subsection{Loop detection} 
\label{sub:loop}

Detecting and representing possible loops of enriched abstract operators 
in an activity schema would help to improve the compactness and to increase 
the applicability of the activity schema. 
In the previous work \cite{vahid2017iros,vahid2017prletter}, we proposed 
a loop detection approach based on the standard methods of computing 
\emph{Suffix Array} ($SA$) of a string --- an array of integers providing 
the starting positions of all suffixes of a string, sorted in lexicographical 
order --- and the \emph{Longest Common Prefix} ($LCP$) array --- an array of 
integers storing the lengths of the longest common prefixes between all 
pairs of consecutive suffixes in a suffix array \cite{manber1993suffix}. 
\footnote{ Suffix array and the longest common prefix array allow 
efficient implementations of many important string operations.}
Since the $LCP$ algorithm also selects the overlapping longest repeated 
substrings in a string, it cannot be independently used to detect potential 
loops in the string. We extend the definition of the $LCP$ to the Non-overlapping 
Longest Common Prefix ($NLCP$) and build an $NLCP$ array from a string:

\begin{definition}%[\textbf{Non-overlapping Longest Common Prefix}]
\label{def:nlcp}
Let $A$ and $B$ be two strings, and $A[i:j]$ and $B[i:j]$ denote the 
substrings of $A$ and $B$ ranging from $i$ to $j-1$ respectively. 
The length of the \emph{Non-overlapping Longest Common Prefix} ($NLCP$) of 
$A$ and $B$, denoted by \ttf{nlcp}$(A,B)$, is the largest integer 
$l \leq \ttf{min}(\Len{A},\Len{B},\allowbreak\ttf{abs}(\Len{A}-\Len{B}))$ such that 
$A[0:l] = B[0:l]$.
\end{definition}

\begin{definition}%[\textbf{Non-overlapping Longest Common Prefix Array}]
\label{def:nlcpa}
Let $S$ be a string and $SA$ the suffix array of $S$. 
% The \emph{Non-over\-lapping Longest Common Prefix} ($NLCP$) array 
An $NLCP$ array, built from $S$ and $SA$, 
is an array of integers of size $n=\ttf{len}(S)$ such that $NLCP[0]$ 
is undefined and $NLCP[i]=\ttf{nlcp}(S[SA[i-1]:n],S[SA[i]:n])$, 
for $1\leq i<n$.
\end{definition}

The $NLCP$ array gives a list of potential patterns in a string, however, 
it does not warrant the obtained patterns are consecutive. We 
proposed the Contiguous Non-overlapping Longest Common Prefix array 
obtained form the $NLCP$ array:

\begin{definition}%[\textbf{Contiguous Non-overlapping Longest Common Prefix Array}]
\label{def:cnlcp}
A \emph{Contiguous Non-overlapping Longest Common Prefix} ($CNLCP$) array 
is an array of structures, constructed from the $SA$ and $NLCP$ arrays of a 
string, such that each $CNLCP[i]$, for $i\geq 0$, contains a substring, 
representing a pattern that consecutively occurs in the string, and a list of 
starting positions of the pattern in the string. A non-overlapping longest 
common prefix between $NLCP[i]$ and $NLCP[i-1]$ is consecutive if 
$NLCP[i]=\ttf{abs}($SA$[i]-$SA$[i-1])$ for $1\leq i<n$. 
\end{definition}

When the $CNLCP$ array is constructed for the abstract plan of a generalized 
and abstracted experience (represented as a string), we start by selecting an 
iteration with the largest length in the $CNLCP$ array and construct a loop 
by merging all iterations of the loop, that is, the loop iterations are merged 
and an intersection of their corresponding features is computed, and a new 
variable represents the different variables playing the same role in the 
corresponding abstract operators and in their corresponding features in each 
subsequence. We continue this process for all iterations in the $CNLCP$ array 
until no more loops are formed. 
In Appendix~\ref{sec:app}, we present an updated version of the $CNLCP$ 
algorithm and a concrete example of computing the CNLCP array.
\lstref{schema_loop} shows a learned activity schema of the `stack' task in 
the \StackDom EBPD with two potential loops of actions.

The specific algorithms for learning activity schemata have been 
described in \cite{mokhtari2016jint,vahid2017prletter,vahid2017iros}.

%%%%%%%%%%%%%%%%%%%%%%%%%%%%%
\begin{lstlisting}[style=customlst,
    label=lst:schema_loop,
    float=t,
    captionpos=b,
    % aboveskip=-5pt,
    % belowskip=-5pt,
    % abovecaptionskip=10pt,
    morekeywords={parameters, domain, objects,
    define, activity, schema, method, abstract, plan, precondition},
    caption={A learned activity schema for the `stack' task with loops 
    (note only key-properties that make distinction between abstract 
    operators are shown and the rest are omitted due to limited space).
    There are two loops in this activity schema. 
    During problem solving, iterations of loops are generated for blue 
    and red blocks on a table, respectively.
    }]
(:method stack
 :parameters    (?table ?pile)
 :abstract-plan ((!pick ?block1 ?table)
                      ((static(blue ?block1))$\cdots$)
                 (!stack ?block1 ?pallet ?pile)
                      ((static(blue ?block1))(static(pallet ?pallet))$\cdots$)
                 $\text{\textbf{(loop}}$ (!pick ?block2 ?table)
                            ((static(blue ?block2))$\cdots$)
                       (!stack ?block2 ?block1 ?pile)
                            ((static(blue ?block1))(static(blue ?block2))$\cdots$)$\text{\textbf{)}}$
                 (!pick ?block5 ?table)
                      ((static(red ?block5))$\cdots$)
                 (!stack ?block5 ?block4 ?pile)
                      ((static(blue ?block4))(static(red ?block5))$\cdots$)
                 $\text{\textbf{(loop}}$ (!pick ?block6 ?table)
                            ((static(red ?block6))$\cdots$)
                       (!stack ?block6 ?block5 ?pile)
                            ((static(red ?block5))(static(red ?block6))$\cdots$)$\text{\textbf{)}}$
                 (!pick ?block8 ?table)
                      ((static(red ?block8))(end(top ?block8 ?pile))$\cdots$)
                 (!stack ?block8 ?block7 ?pile)
                      ((static(red ?block7))(static(red ?block8))$\cdots$)))
\end{lstlisting}
%%%%%%%%%%%%%%%%%%%%%%%%%%%%%

%%%%%%%%%%%%%%%%%%%%%%%%%%%%%%%%%%%%%%%%%%%%%%%%%%%%%%%%%%%
%%%%%%%%%%%%%%%%%%%%%%%%%%%%%%%%%%%%%%%%%%%%%%%%%%%%%%%%%%%
%%%%%%%%%%%%%%%%%%%%%%%%%%%%%%%%%%%%%%%%%%%%%%%%%%%%%%%%%%%
%%%%%%%%%%%%%%%%%%%%%%%%%%%%%%%%%%%%%%%%%%%%%%%%%%%%%%%%%%%
%%%%%%%%%%%%%%%%%%%%%%%%%%%%%%%%%%%%%%%%%%%%%%%%%%%%%%%%%%%
%%%%%%%%%%%%%%%%%%%%%%%%%%%%%%%%%%%%%%%%%%%%%%%%%%%%%%%%%%%
%%%%%%%%%%%%%%%%%%%%%%%%%%%%%%%%%%%%%%%%%%%%%%%%%%%%%%%%%%%

\section{INFERRING THE SCOPE OF AN ACTIVITY SCHEMA}
\label{sec:tvla_learning}

To extend the EBPDs framework, we propose to infer the \emph{scope of 
applicability} of the learned activity schema. The scope allows for testing 
the applicability of the activity schema to solve a given problem. We 
develop an approach based on Canonical Abstraction \citep{SRW:TOPLAS02}, 
which creates a finite representation of a (possibly infinite) set of 
logical structures. The approach is based on Kleene's $3$-valued logic 
\citep{kleene1952introduction}, which extends Boolean logic by introducing 
an indefinite value $\Half$, to denote either $0$ or $1$. 
We infer the scope of an activity schema from the key-properties of a 
generalized and abstracted experience in the form of a $3$-valued logical 
structure, which can be used as an abstraction of a larger $2$-valued 
logical structure. 

We first represent the key-properties of a generalized and abstracted 
experience using a $2$-valued logical structure: 

%%%%%%%%%%%%%%%%%%%%%%%%%%%%%%%%%%%%%%%%%%%%%%%%%%%%%%%%%%%%%%%%%%%%%%%%%%%%%
\begin{definition}%[\textbf{2-Valued Logical Structure}]
\label{def:2_valued}
A \emph{$2$-valued logical structure}, 
also called a \emph{concrete structure}, 
over a set of predicate symbols $P$ and a set of temporal symbols $T$
is a pair, 
\[ C=( U, \iota), \]
where $U$ is a set of individuals called the universe of $C$ 
%(i.e., the set objects involved in $K$) 
and $\iota$ is an interpretation for $P$ and $T$ over $U$. 
The interpretation of a predicate symbol $p\in P$ with 
a temporal symbol $\tau\in T$, denoted by $\iota(\tau(p))$, 
is a function mapping $\tau(p)$ over the universe $U$ to its 
the truth-value in $C$: 
for every predicate symbol $p^{(k)}$ of arity $k$ and temporal symbol 
$\tau$, $\iota(\tau(p)):U^k \to \{0,1\}$.
\end{definition}

% %%%%%%%%%%%%%%%%%%%%%%%%%%%%%%%%%%%%%%%%%%%%%%%%%%%%%%%%%%%%%%%%%%%%%%%%%%%%%
A set of key-properties $K$ is converted into a $2$-valued logical structure, 
denoted by $\Struc(K) =(U, \iota)$, as follows:
\[
\begin{array}{rcl}~ 
P     &=& \bigcup\limits_{\tau(p(t_1,\dots,t_k)) \in K}\{p\}\enspace, \\
T     &=& \bigcup\limits_{\tau(p(t_1,\dots,t_k)) \in K}\{\tau\}\enspace, \\
U     &=& \bigcup\limits_{\tau(p(t_1,\dots,t_k)) \in K}\{t_1,\dots,t_k\}\enspace, \\
\iota &=& \lambda\tau\in T, p^{(k)}\in P~.~\lambda(t_1,\dots,t_k)\in U^k.
          \left\{\begin{array}{ll}
                    1, & \text{if}\quad \tau(p(t_1,\dots,t_k)) \in K\hbox{;} \\
                    0, & \hbox{otherwise.}
                   \end{array}\right.
\end{array}
\]
That is, the universe of $\Struc(K)$ consists of the objects appearing 
in the key-properties of $K$, and the interpretation is defined over the 
key-properties of $K$. The interpretation of a temporal symbol $\tau\in T$, 
where $T=\{\Static,\Init,\End\}$, and a predicate symbol $p^{(k)}\in P$ 
of arity $k$, for a tuple of objects $(t_1,\dots,t_k)\in U$ 
is $1$, i.e., $\iota(\tau(p))(t_1,\dots,t_k)=1$, if a corresponding 
key-property $\tau(p(t_1,\dots,t_k))$ appears in $K$. 

\begin{figure*}[!t]
    \centering
    \begin{subfigure}[b]{\textwidth}
        \includegraphics[width=.97\textwidth]{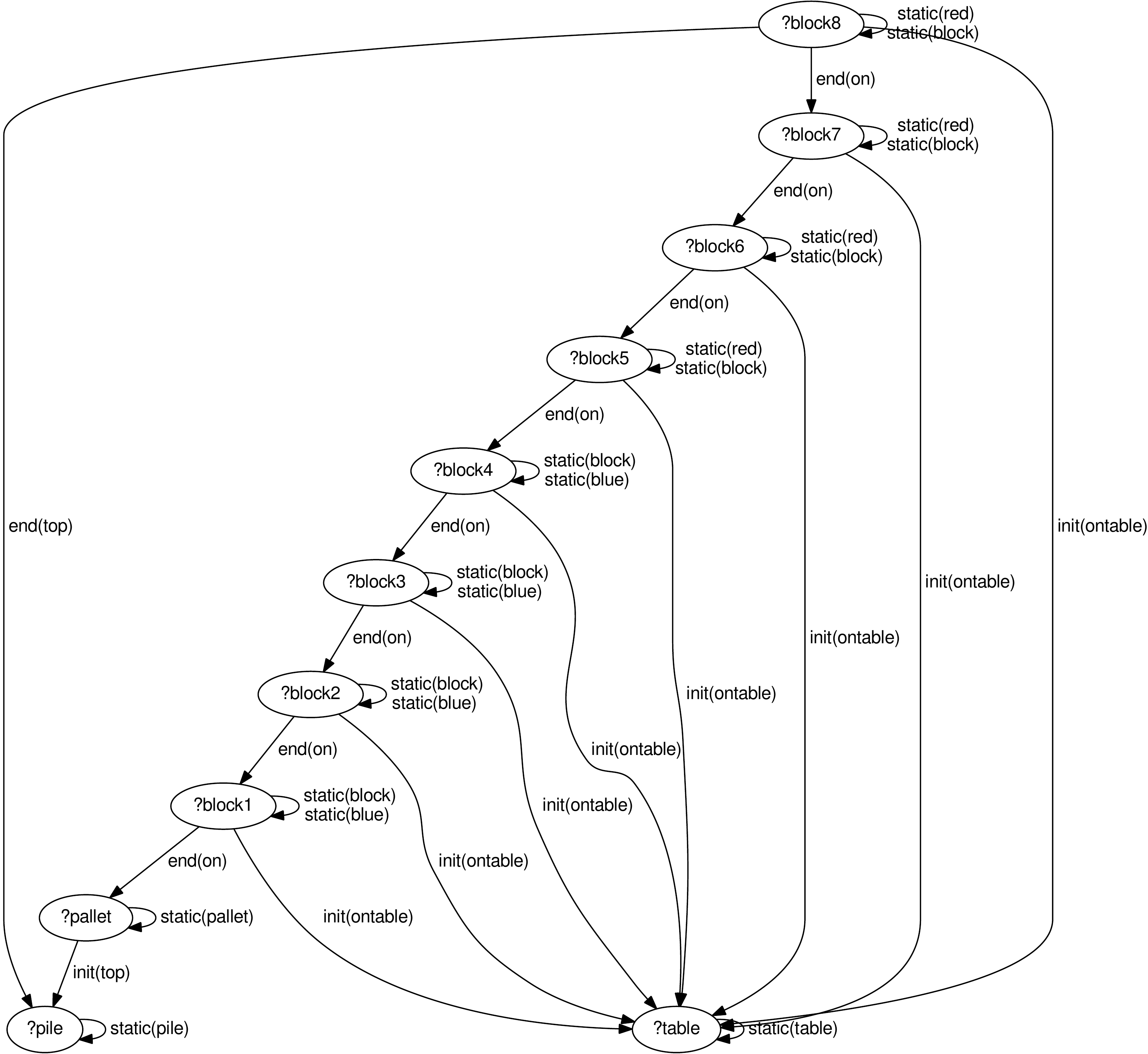}
        \caption{A concrete structure $C$.}
        % \label{fig:concrete_structure}
    \end{subfigure}
    \\\vspace{5pt}
    \begin{subfigure}[b]{.58\textwidth}
        \includegraphics[width=\textwidth]{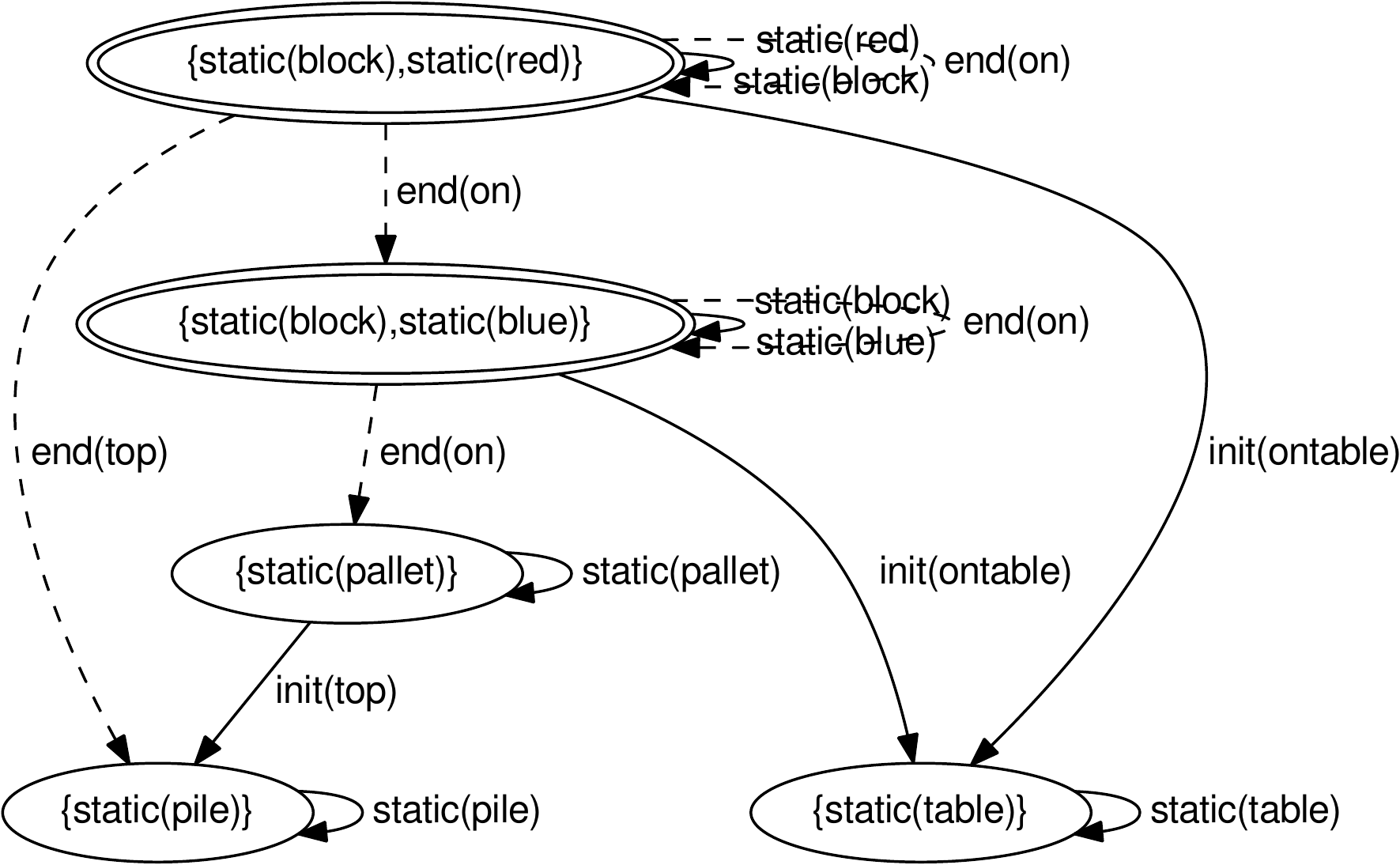}
        \caption{An abstract structure $\Scope=\beta(C)$.}
        % \label{fig:abstract_structure}
    \end{subfigure}
\caption{Canonical abstraction of the (generalized and abstracted) `stack' 
experience (in Listing~\ref{lst:experience_gen}) in the \StackDom 
EBPD. Nodes constitute the universe of a structure and edges represent the 
truth-values of the key-properties over the universe.}
\label{fig:tvla_abstraction}
\vspace{-25pt}
\end{figure*}

$2$-valued logical structures are drawn as directed graphs. The individuals 
of the universe are drawn as nodes, and the key-properties with definite 
values ($1$) are drawn as directed edges. 
% \begin{example}
% \label{ex:2_valued}
For example, \figref{tvla_abstraction}(a) shows a $2$-valued logical 
(concrete) structure $C$ representing the generalized and abstracted 
experience in Listing~\ref{lst:experience_gen}. 
In this example, the universe and the truth-values (interpretations) of 
the key-properties over the universe of $C$ are as follows: 
\footnote{ The truth-value 
of a predicate is $0$ if it is not present in $\iota$.}

\[
\begin{array}{rcl}
P &=& \{ \ttfs{pile}, \ttfs{table}, \ttfs{pallet}, \ttfs{block} \}, \\
T &=& \{ \ttfs{static}, \ttfs{init}, \ttfs{end} \}, \\
U &=& \{ \ttfs{?pile}, \ttfs{?table}, \ttfs{?pallet}, \ttfs{?block1}, 
         \ttfs{?block2}, \ttfs{?block3}, \ttfs{?block4}, \ttfs{?block5}, \\
  & &    \;\ttfs{?block6}, \ttfs{?block7}, \ttfs{?block8} \}\enspace, \\
\iota   &=& \{ \ttfs{(static(pile ?pile))}      = \; 1, \\
        & & \;\ttfs{(static(table ?table))}   = \; 1, \\
        & & \;\ttfs{(static(pallet ?pallet))} = \; 1, \\
        & & \;\ttfs{(static(block ?block1))}  = \; 1, \\
        & & \;\ttfs{(static(block ?block2))}  = \; 1, \\
        & & \;\ttfs{(static(block ?block3))} = \; 1, \\
        & & \;\ttfs{(static(block ?block4))} = \; 1, \\
        & & \;\;\vdots \qquad\}\enspace.
\end{array}
\]
% \end{example}

The scope inference procedure converts a $2$-valued logical structure 
into a $3$-valued logical structure \citep{SRW:TOPLAS02}: 

%%%%%%%%%%%%%%%%%%%%%%%%%%%%%%%%%%%%%%%%%%%%%%%%%%%%%%%%%%%%%%%%%%%%%%%%%%%%%
\begin{definition}%[\textbf{3-Valued Logical Structure}]
\label{def:3_valued}
A \emph{$3$-valued logical structure}, 
also called an \emph{abstract structure}, 
over a set of predicate symbols $P$ and a set of temporal symbols $T$
is a pair, 
\[ \Scope=( U, \iota), \]
where $U$ is a set of individuals called the universe of $\Scope$ 
and $\iota$ is an interpretation $\iota(\tau(p))$ for every predicate 
symbol $p\in P$ and temporal symbol $\tau\in T$. 
For every predicate symbol $p^{(k)}$ of arity $k$ with a temporal symbol 
$\tau$, 
$\iota(\tau(p)):U^k \to \{0,1,\Half\}$, where $\Half$ denotes unknown values. 
\end{definition}

The scope inference procedure converts a $2$-valued logical structure 
into a $3$-valued logical structure \citep{SRW:TOPLAS02}. 
This transformation is based on canonical names, the Kleene's join operation 
\citep{lev2000tvla} and a canonical abstraction function:

%%%%%%%%%%%%%%%%%%%%%%%%%%%%%%%%%%%%%%%%%%%%%%%%%%%%%%%%%%%%%%%%%%%%%%%%%%%%%
\begin{definition}%[\textbf{Canonical Name}]
\label{def:canonical_name}
Let $(U,\iota)$ be a ($2$-valued logical/$3$-valued logical) 
structure over a set of temporal symbols $T$ and 
a set of predicate symbols $P$. 
The \emph{canonical name} of an object $u \in U$, also called 
an \emph{abstraction predicate}, denoted by $\ttf{canon}(u)$, 
is a set of unary predicate symbols with temporal symbols 
that hold for $u$ in the structure: 
\[
\ttf{canon}(u) = \{ \tau(p) \mid \tau\in T, p\in P,\iota(\tau(p))(u)=1\}\enspace.
\]
\end{definition}

%%%%%%%%%%%%%%%%%%%%%%%%%%%%%%%%%%%%%%%%%%%%%%%%%%%%%%%%%%%%%%%%%%%%%%%%%%%%%
For example, the canonical names of the objects in $U$ in the above example
are the following:

\[
\begin{array}{rcl}
\ttfs{canon(?table})  &=& \{\ttfs{static(table)}\}\\
\ttfs{canon(?pile})   &=& \{\ttfs{static(pile)}\}\\
\ttfs{canon(?pallet}) &=& \{\ttfs{static(pallet)}\}\\
\ttfs{canon(?block1}) &=& \{\ttfs{static(block),static(blue)}\}\\
\ttfs{canon(?block2}) &=& \{\ttfs{static(block),static(blue)}\}\\
\ttfs{canon(?block3}) &=& \{\ttfs{static(block),static(blue)}\}\\
\ttfs{canon(?block4}) &=& \{\ttfs{static(block),static(blue)}\}\\
\ttfs{canon(?block5}) &=& \{\ttfs{static(block),static(red)}\}\\
\ttfs{canon(?block6}) &=& \{\ttfs{static(block),static(red)}\}\\
\ttfs{canon(?block7}) &=& \{\ttfs{static(block),static(red)}\}\\
\ttfs{canon(?block8}) &=& \{\ttfs{static(block),static(red)}\}\enspace.
\end{array}
\]
% \end{example}

%%%%%%%%%%%%%%%%%%%%%%%%%%%%%%%%%%%%%%%%%%%%%%%%%%%%%%%%%%%%%%%%%%%%%%%%%%%%%
\begin{definition}%[\textbf{3-Valued Logical Structure}]
\label{def:kleene_join}
In Kleene's $3$-valued logic, let say the values $0$ and $1$ are definite 
values and $\Half$ is an indefinite value. For $l_1,l_2\in\{0, 1, \Half\}$, 
$l_1$ has more definite information than $l_2$, denoted by 
$l_1\preceq l_2$, if $l_1=l_2$ or $l_2=\Half$.
The Kleene's \emph{join operation} of $l_1$ and $l_2$, denoted by 
$l_1\kjoin l_2$, is the least-upper-bound operation with respect to 
$\preceq$ defined as follows:
\[l_1\kjoin l_2 = \left\{\begin{array}{ll}
                     l_1, & \text{if}\quad l_1=l_2\hbox{;} \\
                     \Half, & \hbox{otherwise.}
                    \end{array}
             \right.
\]
\end{definition}

%%%%%%%%%%%%%%%%%%%%%%%%%%%%%%%%%%%%%%%%%%%%%%%%%%%%%%%%%%%%%%%%%%%%%%%%%%%%%%%%
\begin{definition}%[\textbf{Canonical Abstraction}]
\label{def:CanonicalAbstraction}
Let $C=( U, \iota)$ be a $2$-valued logical structure. 
The \emph{canonical abstraction} of $C$, denoted by $\beta(C)$, 
is a $3$-valued logical structure $\Scope=( U', \iota')$ defined as follows:
\begin{align*} 
&U' = \{ \ttf{canon}(u) \mid u \in U \}\enspace,\\
&\iota'(\tau(p^{(k)}))(t_1',\dots,t_k') = 
\KJoin\limits_{t_1,\dots,t_k} \{ \iota(\tau(p^{(k)}))(t_1,\dots,t_k) 
\mid \forall i=1..k.\ t_i' = \ttf{canon}(t_i) \} \enspace.
\end{align*} 

$\Scope$ may contain \emph{summary objects}, that is, 
a set of objects in $U$ with a canonical name, $c$, is 
merged \footnote{Note that we avoid merging objects appearing 
in the task (parameters) of an experience into a summary object.}
into a summary object in $U'$, denoted by \ttf{summary$(c)$}: 
\[
    \ttf{summary}(c)=\{ u\in U \mid \ttf{canon}(u)=c\}\enspace.
\]
\end{definition}

Kleene's join operation determines the truth-value (interpretation) of 
key-pro\-perties in a $3$-valued logical structure. That is, the 
interpretation of a key-property in the $3$-valued logical structure 
is $1$ (solid arrows) if the key-property exists for all objects of 
the same canonical name in the $2$-valued logical structure, otherwise 
the truth-value is $\Half$ if the key-property exists for some objects 
of the same canonical name (dashed arrows), and $0$ if no key-property 
exists.
\footnote{
In a planning domain description, the set of unary predicates is used 
to build the set of abstraction predicates. The function of canonical 
abstraction suggests that we should have sufficient unary predicates 
to be able to determine if an abstract structure exists for a 
concrete structure. In all example domains used in this work, we 
provided sufficient unary predicates. However, the type of objects 
(in typed planning domains descriptions) are also assumed 
as unary predicates when unary predicates are not sufficient.
}

Computing $\beta(\Struc(K))$ for a set of key-properties $K$ of the 
(generalized and abstracted) experience takes polynomial time in 
$|K|+|U|$.

%%%%%%%%%%%%%%%%%%%%%%%%%%%%%%%%%%%%%%%%%%%%%%%%%%%%%%%%%%%%%%%%%%%%%%%%%%%%%
$3$-valued logical structures are also drawn as directed graphs. 
Summary objects are drawn as double circles. Definite values are drawn 
as in the $2$-values logical structures, and indefinite values ($\Half$) 
are drawn as dashed directed edges. 
For example, \figref{tvla_abstraction}(b) shows a $3$-valued logical structure 
$\Scope$ of the concrete structure $C$ in \figref{tvla_abstraction}(a). 
The double circles stand for summary objects, e.g., 
$\ttf{summary}(\allowbreak\{\ttf{(during,block),(during,blue)}\})$ is a 
summary object in $\Scope$ corresponding to the objects 
(\texttt{?block1..?block4}) in $C$ with the same canonical name. 
Solid (dashed) arrows represent truth-values of $1$ ($\Half$). 
Intuitively, because of the summary objects, the abstract structure $\Scope$ 
represents the concrete structure $C$ and all other `Stack' problems that 
have exactly one \ttf{table}, one \ttf{pile}, one \ttf{pallet}, and at 
least one \ttf{blue block} and one \ttf{red block} such that the blocks 
are initially on the table and finally red blocks are on top of blue blocks 
in the pile. 

%%%%%%%%%%%%%%%%%%%%%%%%%%%%%%%%%%%%%%%%%%%%%%%%%%%%%%%%%%
%%%%%%%%%%%%%%%%%%%%%%%%%%%%%%%%%%%%%%%%%%%%%%%%%%%%%%%%%%

The inferred scope is finally represented in a learned activity schema 
as a set of key-properties. 
Summary objects are represented as \ttf{(summary ?c)} where \ttf{?c} is a 
canonical name. Indefinite values ($\Half$) appear as 
\texttt{(maybe($p$))} where $p$ represents a key-property. 
\lstref{schema_loop_prec} shows the inferred scope of applicability 
for the activity schema of the `stack' task. 

%%%%%%%%%%%%%%%%%%%%%%%%%%%%%
\begin{figure}[!t]
\lstinputlisting[style=customlst,
    label=lst:schema_loop_prec,
    % float,
    captionpos=b,
    % numbersep=-4.5pt,
    %breaklines=true,
    % abovecaptionskip=5pt,
    morekeywords={parameters, domain, objects,
    define, activity, schema, method, abstract, plan, scope},
    % caption={A learned activity schema for the `stack' 
    % task with loops of actions and its scope of applicability.
    caption={The inferred scope of applicability for the learned 
    activity schema of the `stack' task.
    `Summary' objects represent arbitrary numbers of objects of the same 
    canonical name, and `maybe' key-properties represent key-properties 
    with truth-values of either $0$ or $1$ in a task planning problem.
    }]
% {listings/robotic_arm_schema_loop_prec.ebpd}
{listings/scope}
\end{figure}
%%%%%%%%%%%%%%%%%%%%%%%%%%%%%

%%%%%%%%%%%%%%%%%%%%%%%%%%%%%%%%%%%%%%%%%%%%%%%%%%%%%%%%%%%
%%%%%%%%%%%%%%%%%%%%%%%%%%%%%%%%%%%%%%%%%%%%%%%%%%%%%%%%%%%
%%%%%%%%%%%%%%%%%%%%%%%%%%%%%%%%%%%%%%%%%%%%%%%%%%%%%%%%%%%
%%%%%%%%%%%%%%%%%%%%%%%%%%%%%%%%%%%%%%%%%%%%%%%%%%%%%%%%%%%
%%%%%%%%%%%%%%%%%%%%%%%%%%%%%%%%%%%%%%%%%%%%%%%%%%%%%%%%%%%
%%%%%%%%%%%%%%%%%%%%%%%%%%%%%%%%%%%%%%%%%%%%%%%%%%%%%%%%%%%
%%%%%%%%%%%%%%%%%%%%%%%%%%%%%%%%%%%%%%%%%%%%%%%%%%%%%%%%%%%

\section{SELECTING AN APPLICABLE ACTIVITY SCHEMA FOR PROBLEM SOLVING}
\label{sec:tvla_execution}

When an activity schema is learned for a class of problems, it can be used 
to generate a solution plan for a given task problem. 
In the previous work, the EBPDs framework lacked an automatic strategy to 
find an applicable activity schema, among several learned activity schemata, 
for solving a task problem.
Here, we extend the previous work in which an activity schema is selected 
as applicable to solving a given task problem if the task problem is 
\emph{embedded} in the scope of the activity schema, i.e., the task 
problem maps onto the scope of the activity schema. 
Selecting an activity schema involves \emph{problem abstraction} and 
\emph{testing the scope of applicability} (i.e., embedding). 

Given the predicate abstraction hierarchy in $\mc{R}$, the abstraction of a 
problem is achieved by transforming the concrete predicates into abstract 
predicates, which results in an abstracted task problem. 
A concrete task problem $\mc{P}=( t,\sigma,s_0,g )$ is translated into an 
abstracted task problem $\mc{P}_a=( t,\sigma_a,{s_0}_a,g_a )$, denoted by 
$\ttf{Abs}(\mc{P})$, as follows:
\[
\sigma_a = \{\ttf{parent}(p) \mid p\in\sigma\}, \quad
{s_0}_a  = \{\ttf{parent}(p) \mid p\in s_0\}, \quad
g_a      = \{\ttf{parent}(p) \mid p\in g\}.
\]

To see how the abstracted task problem $\mc{P}_a$ is embedded in the scope of 
an activity schema, we first convert $\mc{P}_a$ into a $2$-valued structure 
\footnote{ More precisely, to represent an abstracted task problem 
$\mc{P}=(t,\sigma,{s_0},g)$ into a $2$-valued structure, we generate 
a set of key-properties $K$ for $\mc{P}$ by wrapping the predicates 
of $(\sigma,{s_0},g)$ with temporal symbols \ttf{static}, \ttf{init} 
and \ttf{end}, and then convert $K$ into a $2$-valued structure.
} 
(as described in Section~\ref{sec:tvla_learning}), 
and then test if the obtained $2$-valued structure is embedded in the 
scope of an activity schema:

\begin{definition}%[\textbf{Embedding}]
\label{def:embedding}
We say that a concrete structure (i.e., an abstracted task problem represented 
in a $2$-valued logical structure) $C=(U, \iota)$ is \emph{embedded} in an 
abstract structure (i.e., in the scope of an activity schema) $\Scope=(U', \iota')$, 
denoted by $C \sqsubseteq \Scope$, if there exists a function $f:U \to U'$ 
such that $f$ is surjective and 
for every predicate symbol $p^{(k)}$ of arity $k$ with a temporal symbol $\tau$, 
and tuple of objects $u_1,...,u_k \in U$, one of the following conditions holds:
\begin{equation}
\label{eq:embedding}
\begin{array}{c}
\iota(\tau(p))(u_1,...,u_k) = \iota'(\tau(p))(f(u_1),...,f(u_k))
\quad\text{or}\quad
\iota'(\tau(p))(f(u_1),...,f(u_k)) = \Half \enspace.
\end{array}
\end{equation}

% \noindent
Further, $\Scope$ represents the set of concrete 
structures embedded in it: $\{C \mid C \sqsubseteq \Scope\}$.
\end{definition}

\begin{proposition}
Canonical abstraction is sound  with respect to the embedding relation.
That is, $C \sqsubseteq \beta(C)$ holds for every concrete structure $C$.
\end{proposition}

\begin{proposition}
If an abstract structure $\Scope=(U', \iota')$
is in the image of canonical abstraction, then
checking whether a concrete structure
$C=(U, \iota)$ is embedded in $\Scope$ 
can be done in time polynomial in $|U|+|U'|+|K|$.

\end{proposition}

\noindent\textbf{\textit{Proof sketch.}}
Observe that \eqref{eq:embedding} implies, if $C$ is embedded in $\Scope$,
then $\Scope$ and $C$ have equal sets of canonical names (checkable in 
polynomial time). 
Therefore, the embedding function must be
$f = \{ u \mapsto u' \mid  \ttf{canon}_C(u)=\ttf{canon}_{\Scope}(u')\}$.
Checking that \eqref{eq:embedding} holds for $f$ takes polynomial time.
\qed

Based on the above definition, we implemented and integrated an 
\textsc{Embedding} function into the EBPDs system to find an 
applicable activity schema $m=( h,\Scope,\Omega )$ 
to a task planning problem $\mc{P}=(t,\sigma,s_0,g)$, 
by checking whether $\Struc(\ttf{Abs}(\mc{P})) \sqsubseteq \Scope$ holds.

%%%%%%%%%%%%%%%%%%%%%%%%%%%%%%%%%%%%%%%%%%%%%%%%%%%%%%%%%%%
%%%%%%%%%%%%%%%%%%%%%%%%%%%%%%%%%%%%%%%%%%%%%%%%%%%%%%%%%%%
%%%%%%%%%%%%%%%%%%%%%%%%%%%%%%%%%%%%%%%%%%%%%%%%%%%%%%%%%%%
%%%%%%%%%%%%%%%%%%%%%%%%%%%%%%%%%%%%%%%%%%%%%%%%%%%%%%%%%%%
%%%%%%%%%%%%%%%%%%%%%%%%%%%%%%%%%%%%%%%%%%%%%%%%%%%%%%%%%%%
%%%%%%%%%%%%%%%%%%%%%%%%%%%%%%%%%%%%%%%%%%%%%%%%%%%%%%%%%%%
%%%%%%%%%%%%%%%%%%%%%%%%%%%%%%%%%%%%%%%%%%%%%%%%%%%%%%%%%%%

%%%%%%%%%%%%%%%%%%%%%%%%%%%%%%%%%%%%%%%%%%%%%%%%%%%%%%%%%%%%%%%%%%%%%%%%
\section{PLANNING USING THE LEARNED ACTIVITY SCHEMATA}
\label{sec:planner}

We have previously proposed a planning system for generating a solution 
plan to a given task problem using a learned activity schema 
\cite{vahid2017iros,vahid2017prletter}. 
Problem solving in EBPDs is achieved using a hierarchical problem solver 
which includes an abstract planner---\textit{Abstract Schema-Based Planner} 
(ASBP), and a concrete planner---\textit{Schema-Based Planner} (SBP).
Given an experience-based planning domain $\EBPD$ and a task planning 
problem $\mc{P}$, the EBPDs' planning system retrieves an applicable 
activity schema $\Sch$, i.e., checks 
$\Struc(\ttf{Abs}(\mc{P})) \sqsubseteq \Scope$ 
(see Section~\ref{sec:tvla_execution}), and attempts 
to generate a solution plan to $\mc{P}$. 

Using the abstract planning domain $\mc{D}_a$, ASBP first 
derives an abstract solution by instantiating the enriched abstract plan 
$\Omega$ for $\ttf{Abs}(\mc{P})$. This also involves extending possible 
loops in $\Omega$ for the applicable objects in $\ttf{Abs}(\mc{P})$. 
To extend a loop, ASBP simultaneously generates all successors for an 
iteration of the loop and for the following enriched abstract operator 
after the loop. 
ASBP computes a cost for all generated successors based on the number of 
features of abstract operators verified with the features extracted for the 
instantiated abstract actions, and selects the best current action with the 
lowest cost during the search. 
Finally, a ground abstract plan, $\Pi$, is generated when ASBP gets the end 
of (the enriched abstract plan) $\Omega$ (where the goal must also be achieved). 

The ground abstract plan $\Pi$ produced by ASBP becomes the main skeleton 
of the final solution based on which SBP, using the concrete planning domain 
$\mc{D}_c$, produces a final solution plan to $\mc{P}$ by generating and 
substituting concrete actions for the abstract actions in $\Pi$ 
(as specified in the operator abstraction hierarchy in $\mc{R}$). 
This might also involve generating and inserting actions from the $\varnothing$ 
($nil$) class (see Table~\ref{tbl:operator_hierarchies}). 

The specific planning algorithm and its respective implementation have been 
proposed and described in \cite{vahid2017iros,vahid2017prletter}. 

%%%%%%%%%%%%%%%%%%%%%%%%%%%%%%%%%%%%%%%%%%%%%%%%%%%%%%%%%%%
%%%%%%%%%%%%%%%%%%%%%%%%%%%%%%%%%%%%%%%%%%%%%%%%%%%%%%%%%%%
%%%%%%%%%%%%%%%%%%%%%%%%%%%%%%%%%%%%%%%%%%%%%%%%%%%%%%%%%%%
%%%%%%%%%%%%%%%%%%%%%%%%%%%%%%%%%%%%%%%%%%%%%%%%%%%%%%%%%%%
%%%%%%%%%%%%%%%%%%%%%%%%%%%%%%%%%%%%%%%%%%%%%%%%%%%%%%%%%%%
%%%%%%%%%%%%%%%%%%%%%%%%%%%%%%%%%%%%%%%%%%%%%%%%%%%%%%%%%%%
%%%%%%%%%%%%%%%%%%%%%%%%%%%%%%%%%%%%%%%%%%%%%%%%%%%%%%%%%%%

%%%%%%%%%%%%%%%%%%%%%%%%%%%%%%%%%%%%%%%%%%%%%%%%%%%%%%%%%%%
\section{EMPIRICAL EVALUATION}\label{sec:experiments}

We present the results of our experiments in different classes of problems. 

%%%%%%%%%%%%%%%%%%%%%%%%%%%%%%%%%%%%%%%%%%%%%%%%%%%%%%%%%%%%%%%%%%%%%%%%%%%%%%%
\subsection{Prototyping and implementation}
\label{sec:implementation}

We implemented a prototype of our system in \textsc{SWI-Prolog}, 
which is a general-purpose logic programming language for fast 
prototyping artificial intelligence techniques, and used TVLA 
as an engine for computing the scope of applicability of activity 
schemata. We performed all experiments in simulated domains and 
simulated robot platforms, e.g., PR2, on a machine $2.20$GHz Intel 
Core i$7$ with $12$G memory. 

\subsection{\textsc{STACKING-BLOCKS}}\label{sec:experiments:sim}

In our first experiment, we use the \StackDom EBPD, as described 
in Section~\ref{sec:running_example} (see also Tables~\ref{tbl:predicate_hierarchies} 
and ~\ref{tbl:operator_hierarchies}). 
The main objective of this experiment is to learn a set of different activity 
schemata (tasks) with the same goal but different scopes of applicability, and 
to evaluate how the scope testing (embedding) function allows the system 
to automatically find an applicable activity schema to a given task problem. 
In this paper, we described a class of `stack' problems (Section~\ref{sec:running_example})
with an experience (Listing~\ref{lst:experience}), a learned activity schema 
(Listing~\ref{lst:schema_loop}), and its scope of applicability 
(Listing~\ref{lst:schema_loop_prec} and Figure~\ref{fig:tvla_abstraction}(b)). 

Additionally, we define three other classes of the `stack' problems with the same 
goal but different initial configurations as follow: (\emph{i}) a pile of red and 
blue blocks, with red blocks at the bottom and blue blocks on top; (\emph{ii}) a 
pile of alternating red and blue blocks, with a blue block at the bottom and a red 
block on top; and (\emph{iii}) a pile of alternating red and blue blocks, with a 
red block at the bottom and a blue block on top. 
In all classes of problems, the goal is to make a new pile of red and blue blocks 
with blue blocks at the bottom and red blocks on top (the same goal as in 
Section~\ref{sec:running_example}). 

\begin{figure*}[!t]
    \centering
    \begin{subfigure}[b]{\textwidth}
        \centering
        \captionsetup{width=\textwidth}%
        \includegraphics[width=.72\textwidth]{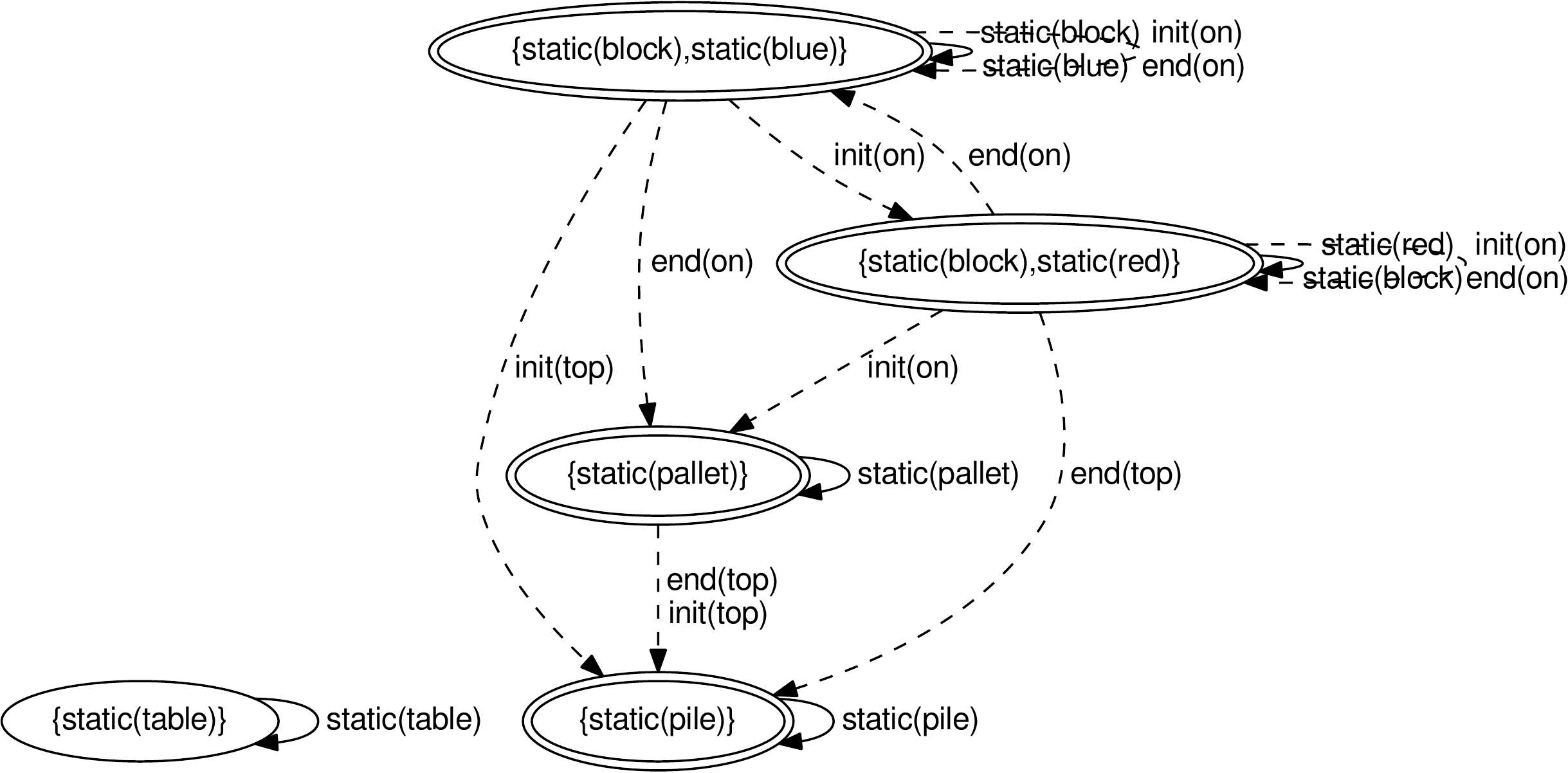}
        \caption{This scope of applicability (abstract structure) represents all 
        `stack' problems that have exactly one \ttf{table} and at least one \ttf{pile}, 
        one \ttf{pallet}, one \ttf{blue block} and one \ttf{red block} such that 
        blue blocks are initially on top of red blocks and finally red blocks are 
        on top of blue blocks (on a pallet) on a pile.}
        \label{fig:abstract_stack3}
    \end{subfigure}
    \\\vspace{2pt}
    \begin{subfigure}[b]{\textwidth}
        \centering
        \captionsetup{width=\textwidth}%
        \includegraphics[width=.72\textwidth]{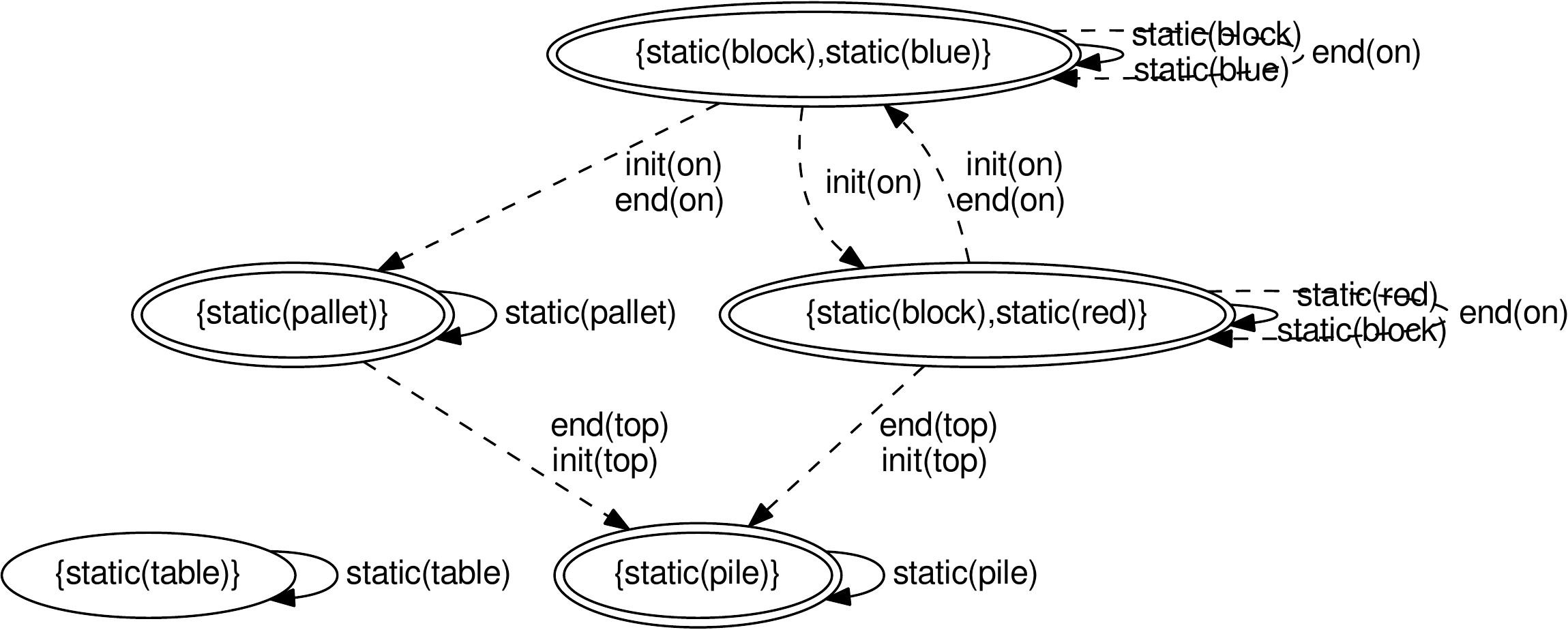}
        \caption{This scope of applicability represents all `stack' problems that 
        have exactly one \ttf{table} and at least one \ttf{pile}, one \ttf{pallet}, 
        one \ttf{blue block} and one \ttf{red block} such that alternate red and blue 
        blocks are initially on a pile with a blue block at the bottom (on a pallet) 
        and a red block on top and finally red blocks are on top of blue blocks.}
        \label{fig:abstract_stack4}
    \end{subfigure}
    \\\vspace{2pt}
    \begin{subfigure}[b]{\textwidth}
        \centering
        \captionsetup{width=\textwidth}%
        \includegraphics[width=.72\textwidth]{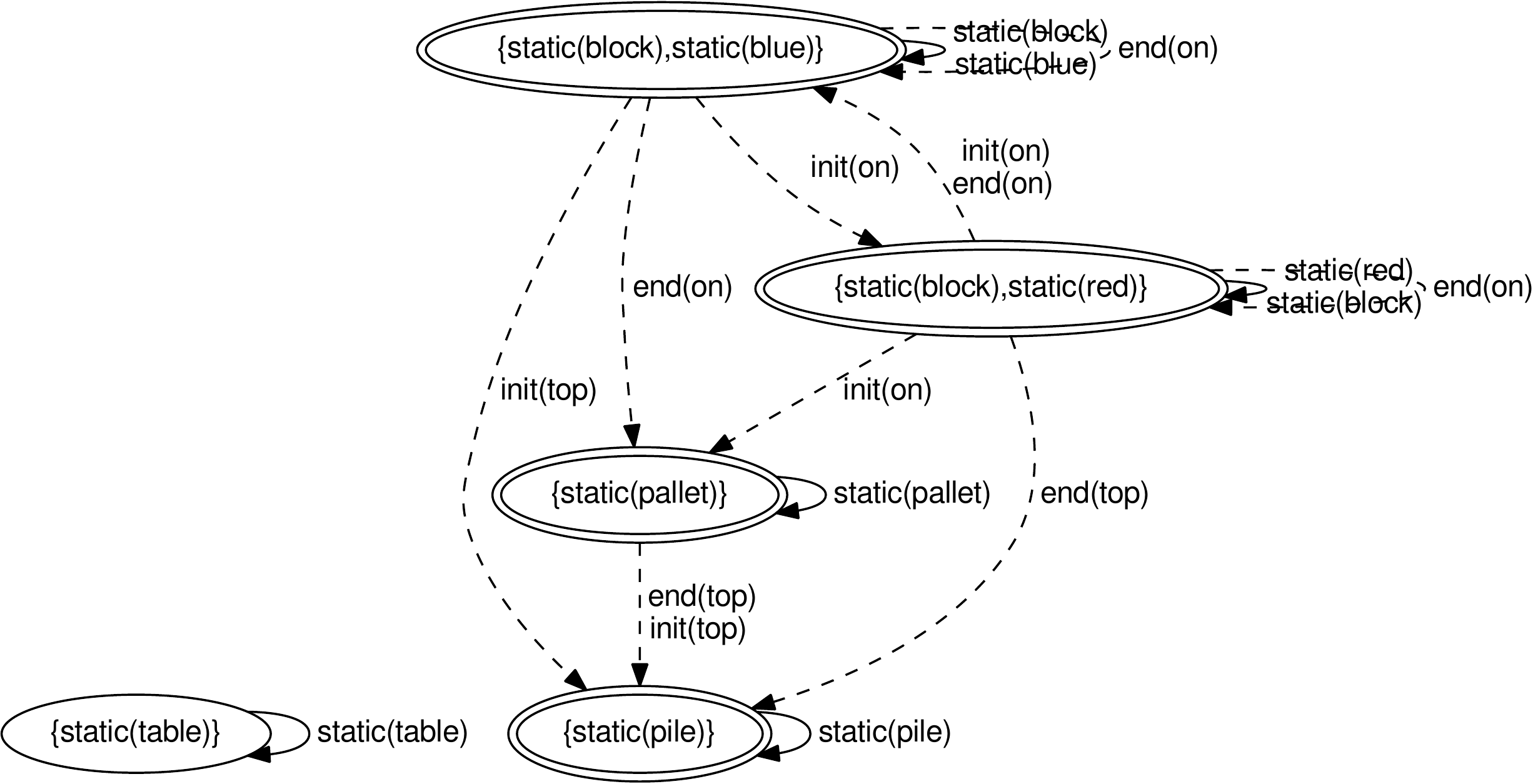}
        \caption{This scope of applicability represents all `stack' problems that 
        have exactly one \ttf{table} and at least one \ttf{pile}, one \ttf{pallet}, 
        one \ttf{blue block} and one \ttf{red block} such that alternate red and blue 
        blocks are initially on a pile with a red block at the bottom (on a pallet) 
        and a blue block on top and finally red blocks are on top of blue blocks.}
        \label{fig:abstract_stack5}
    \end{subfigure}
    \caption{The scope of applicability, i.e., canonical abstraction, of the 
    additional three classes of the `stack' task in the \StackDom EBPD.}
    \label{fig:stack_scopes}
    \vspace{-25pt}
\end{figure*}

To show the effectiveness of the proposed scope of applicability inference, we 
simulated an experience (containing an equal number of $20$ blocks of red and 
blue colors) in each of above classes, which based on them the system generates 
three activity schemata with distinct scopes of applicability 
(see Figure~\ref{fig:stack_scopes}). 

\begin{figure}[!t]
  \centering
  \includegraphics[width=\linewidth]{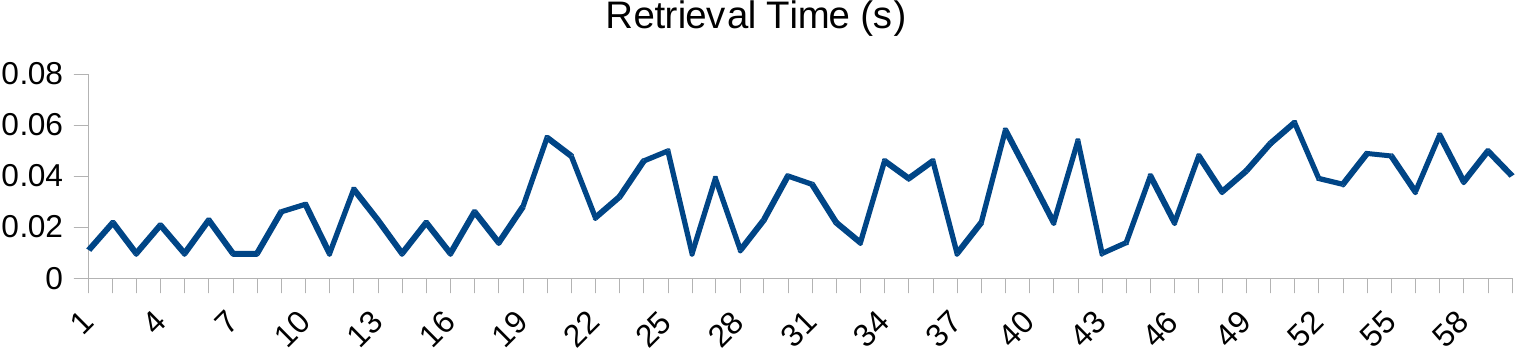}
%  \vspace{-15pt}
  \caption{CPU time used by SBP to find an applicable activity 
  schema (among 4) for solving problems in the \StackDom domain.}
  \label{fig:stack_retrieval}
 \vspace{-5pt}
\end{figure}

\begin{figure}[!t]
  \centering
  \begin{subfigure}[b]{.496\linewidth}
    \includegraphics[width=\linewidth]{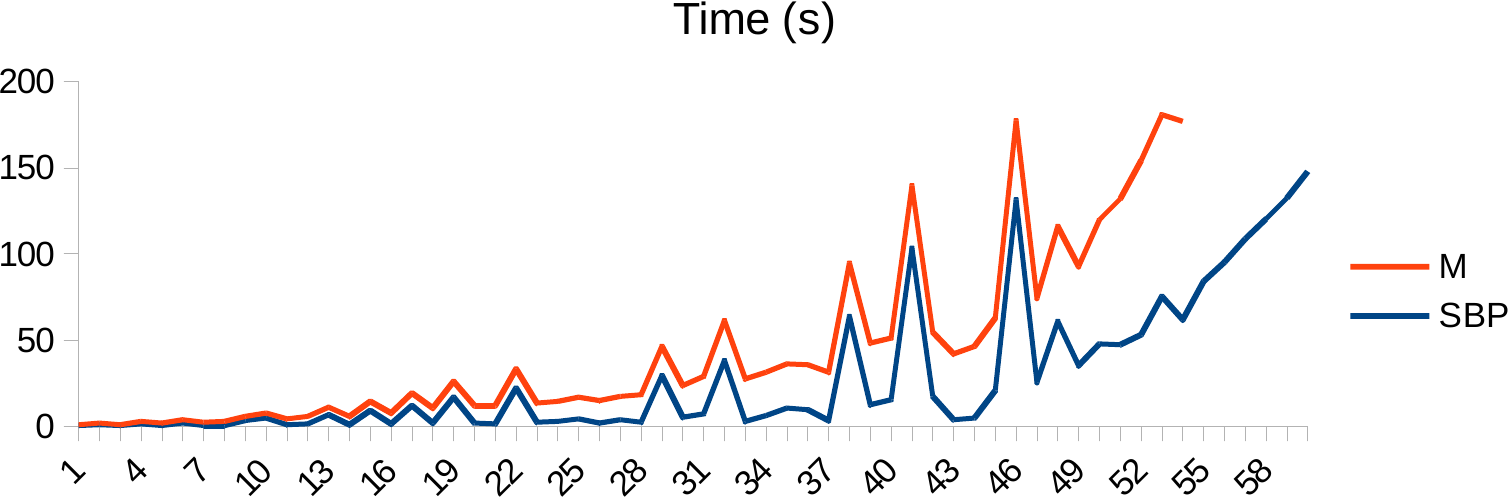}
  \end{subfigure}\hspace{-1.4pt}
  \begin{subfigure}[b]{.496\linewidth}
    \includegraphics[width=\linewidth]{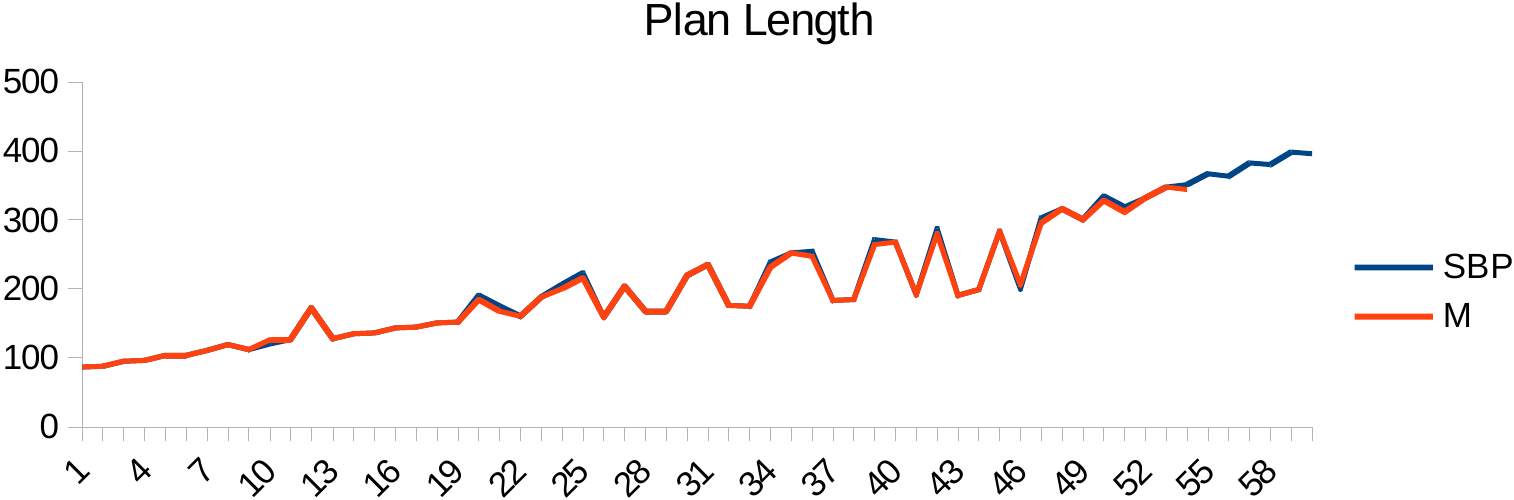}
  \end{subfigure}\\\vspace{5pt}
  \begin{subfigure}[b]{.496\linewidth}
    \includegraphics[width=\linewidth]{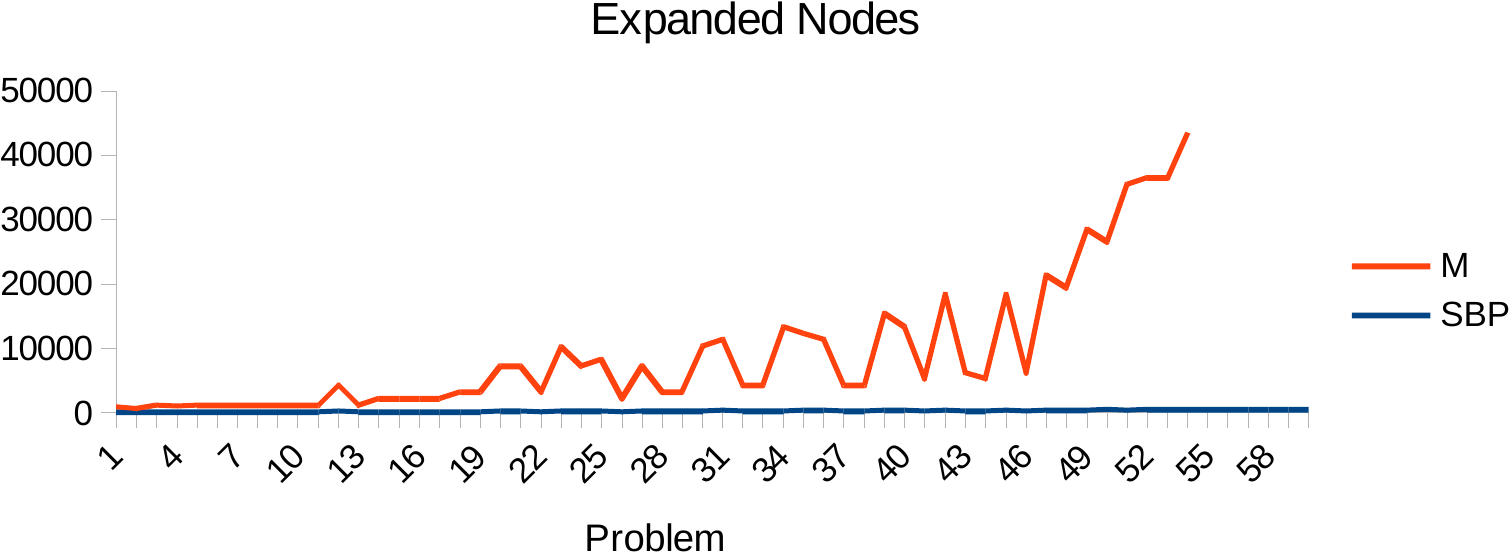}
  \end{subfigure}\hspace{-1.4pt}
  \begin{subfigure}[b]{.496\linewidth}
    \includegraphics[width=\linewidth]{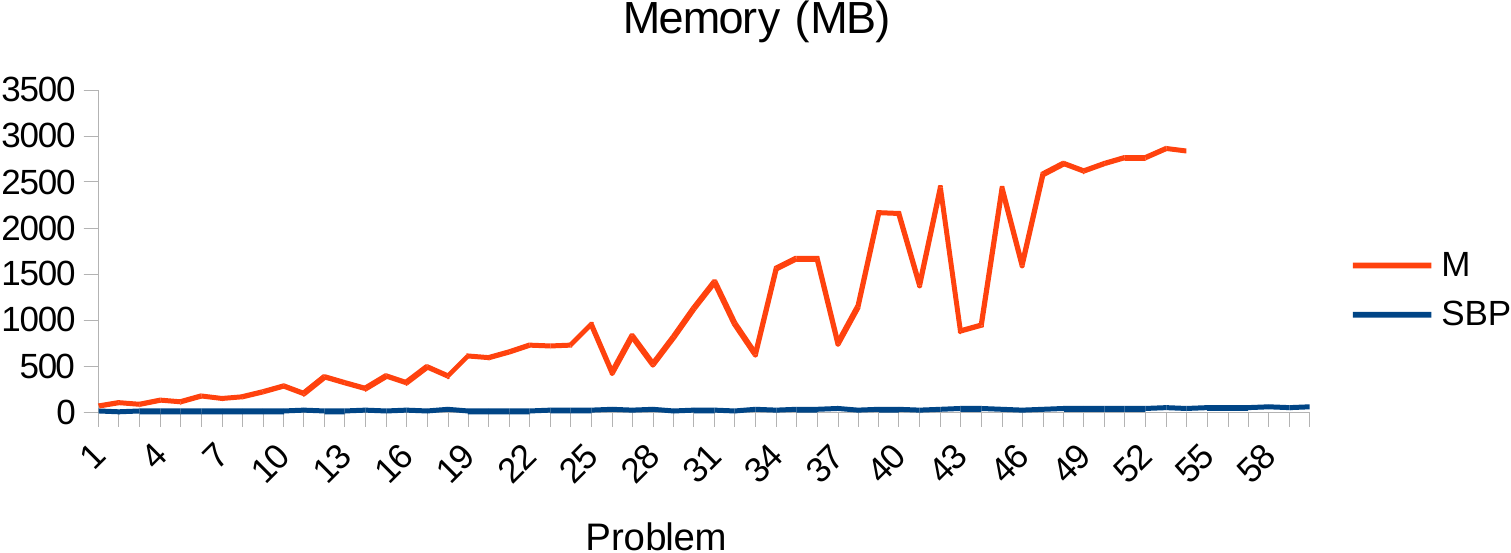}
  \end{subfigure}\\
%  \vspace{-5pt}
  \caption{Performance of the SBP and \textsc{Madagascar} (M) 
  in the \StackDom domain.}
  % \vspace{15pt}
  \label{fig:stack_chart}
 \vspace{-5pt}
\end{figure}

To evaluate the system over the learned activity schemata, we randomly generated 
$60$ task problems in all four classes of the `stack' tasks ($15$ in each class), 
ranging from $20$ to $50$ equal number of red and blue blocks in each problem. 
In this experiment, the system found an applicable activity schema (among $4$) 
to solve given task problems in under $60\text{ms}$ for testing the scope of 
applicability (see Figure~\ref{fig:stack_retrieval}), and then successfully 
solved all problems. 
To show the efficiency of the system, we also evaluated and compared the performance 
of the SBP with a state-of-the-art planner, \textsc{Madagascar} \cite{RINTANEN201245}, 
based on four measures: time, memory, number of evaluated nodes and plan length 
(see Table~\ref{tbl:stack_exp}). 
In this experiment, SBP was extremely efficient in terms of memory and evaluated 
nodes in the search tree. SBP was also fairly fast to solve some problems 
comparing to \textsc{Madagascar}.
Note that the time comparison is not accurate in this evaluation, since SBP has 
been implemented in \textsc{Prolog}, but, by contrast, \textsc{Madagascar} has 
been implemented in \textsc{C++}. 
Figure~\ref{fig:stack_chart} alternatively summarizes the performance of the 
two planners.

%%%%%%%%%%%%%%%%%%%%%%%%%%%%%%%%%%%%%%%%%%%%%%%%%%%%%%%%%%
%%%%%%%%%%%%%%%%%%%%%%%%%%%%%%%%%%%%%%%%%%%%%%%%%%%%%%%%%%

%---------------------------------------------------------------------------
%---------------------------------------------------------------------------
\begin{table}[t!]
\setlength{\tabcolsep}{8pt}
  \centering
  \caption{Performance of the SBP and \textsc{Madagascar} (M) planners in terms of applicability 
  test (retrieval) time, search time, memory, developed nodes and plan length in the 
  different classes of `stack' problems in the \StackDom EBPD.}
  %(sorted by the search time of M)
  % \resizebox{\linewidth}{!}{
  \resizebox*{!}{\dimexpr\textheight-5.5\baselineskip\relax}{
  \begin{tabular}{cccccccccc}
    \hline
    \multirow{1}{*}{Problem/} &
      \multicolumn{1}{c}{Retrieval$^*$ time (s)} &
      \multicolumn{2}{c}{Search time (s)} &
      \multicolumn{2}{c}{Memory (MB)} &
      \multicolumn{2}{c}{Evaluated states} &
      \multicolumn{2}{c}{Plan length} \\
    (\#blocks) & SBP & SBP & M & SBP & M & SBP & M & SBP & M \\
    \hline
p1\ \ \ \ (22) & 0.011 & 0.29   & 0.550   & 10.6 & 57.2    & 131 & 813  & 87  & 87  \\ \hline %% p22 (1)
p2\ \ \ \ (22) & 0.022 & 0.90   & 0.820   & 8.1  & 92.5    & 133 & 597  & 88  & 88  \\ \hline %% p22 (3)
p3\ \ \ \ (24) & 0.010 & 0.37   & 0.820   & 12.4 & 76.9    & 143 & 1011 & 95  & 95  \\ \hline %% p24 (1)
p4\ \ \ \ (24) & 0.021 & 1.44   & 1.250   & 8.9  & 124.9   & 145 & 985  & 96  & 96  \\ \hline %% p24 (3)
p5\ \ \ \ (26) & 0.010 & 0.49   & 1.290   & 13.9 & 100.2   & 155 & 1k   & 103 & 103 \\ \hline %% p26 (1)
p6\ \ \ \ (26) & 0.023 & 2.16   & 1.780   & 9.8  & 162.4   & 157 & 1k   & 104 & 104 \\ \hline %% p26 (3)
p7\ \ \ \ (28) & 0.010 & 0.61   & 1.780   & 15.9 & 130.2   & 167 & 1k   & 111 & 111 \\ \hline %% p28 (1)
p8\ \ \ \ (30) & 0.010 & 0.77   & 2.360   & 17.3 & 148.8   & 179 & 1k   & 119 & 119 \\ \hline %% p30 (1)
p9\ \ \ \ (28) & 0.026 & 3.22   & 2.750   & 10.3 & 212.9   & 169 & 1k   & 112 & 112 \\ \hline %% p28 (3)
p10 \ (30) & 0.029 & 4.67   & 3.170   & 11.4 & 271.2   & 181 & 1k   & 120 & 127 \\ \hline %% p30 (3)
p11 \ (32) & 0.010 & 0.95   & 3.330   & 19.7 & 187.10  & 191 & 1k   & 127 & 127 \\ \hline %% p32 (1)
p12 \ (22) & 0.035 & 1.48   & 4.220   & 16.9 & 369.7   & 271 & 4k   & 172 & 172 \\ \hline %% p22 (4)
p13 \ (32) & 0.023 & 6.84   & 4.460   & 12.5 & 307.4   & 193 & 1k   & 128 & 128 \\ \hline %% p32 (3)
p14 \ (34) & 0.010 & 1.16   & 4.570   & 22.1 & 238.5   & 203 & 2k   & 135 & 135 \\ \hline %% p34 (1)
p15 \ (34) & 0.022 & 9.14   & 5.270   & 13.5 & 382.8   & 205 & 2k   & 136 & 136 \\ \hline %% p34 (3)
p16 \ (36) & 0.010 & 1.43   & 6.390   & 24.1 & 296.7   & 215 & 2k   & 143 & 143 \\ \hline %% p36 (1)
p17 \ (36) & 0.026 & 12.31  & 6.900   & 13.9 & 478.5   & 217 & 2k   & 144 & 144 \\ \hline %% p36 (3)
p18 \ (38) & 0.014 & 1.73   & 8.770   & 26.9 & 366.7   & 227 & 3k   & 151 & 151 \\ \hline %% p38 (1)
p19 \ (38) & 0.028 & 16.87  & 9.200   & 15.0 & 592.2   & 229 & 3k   & 152 & 152 \\ \hline %% p38 (3)
p20 \ (24) & 0.055 & 2.05   & 9.950   & 17.7 & 577.7   & 288 & 7k   & 191 & 184 \\ \hline %% p24 (5)
p21 \ (22) & 0.048 & 1.33   & 10.930  & 16.1 & 642.6   & 264 & 7k   & 175 & 168 \\ \hline %% p22 (5)
p22 \ (40) & 0.024 & 22.23  & 11.100  & 16.1 & 714.4   & 241 & 3k   & 160 & 160 \\ \hline %% p40 (3)
p23 \ (24) & 0.032 & 2.30   & 11.190  & 19.4 & 698.8   & 296 & 10k  & 188 & 188 \\ \hline %% p24 (4)
p24 \ (26) & 0.046 & 3.04   & 11.630  & 20.1 & 709.5   & 312 & 7k   & 207 & 200 \\ \hline %% p26 (5)
p25 \ (28) & 0.050 & 4.50   & 12.560  & 22.5 & 929.2   & 336 & 8k   & 223 & 216 \\ \hline %% p28 (5)
p26 \ (40) & 0.010 & 2.07   & 12.900  & 29.7 & 401.1   & 239 & 2k   & 159 & 159 \\ \hline %% p40 (1)
p27 \ (26) & 0.039 & 3.78   & 13.530  & 21.9 & 802.6   & 321 & 7k   & 204 & 204 \\ \hline %% p26 (4)
p28 \ (42) & 0.011 & 2.48   & 16.120  & 31.5 & 491.7   & 251 & 3k   & 167 & 167 \\ \hline %% p42 (1)
p29 \ (42) & 0.023 & 29.52  & 16.930  & 17.3 & 799.7   & 253 & 3k   & 168 & 168 \\ \hline %% p42 (3)
p30 \ (28) & 0.040 & 5.27   & 18.330  & 23.7 & 1098.0  & 346 & 10k  & 220 & 220 \\ \hline %% p28 (4)
p31 \ (30) & 0.037 & 7.50   & 21.680  & 26.6 & 1388.7  & 371 & 11k  & 236 & 236 \\ \hline %% p30 (4)
p32 \ (44) & 0.022 & 38.26  & 23.090  & 17.9 & 947.1   & 265 & 4k   & 176 & 176 \\ \hline %% p44 (3)
p33 \ (44) & 0.014 & 2.96   & 24.690  & 35.0 & 593.8   & 263 & 4k   & 175 & 175 \\ \hline %% p44 (1)
p34 \ (30) & 0.046 & 6.50   & 24.980  & 24.1 & 1537.7  & 360 & 13k  & 239 & 232 \\ \hline %% p30 (5)
p35 \ (32) & 0.039 & 10.69  & 25.570  & 28.5 & 1645.8  & 396 & 12k  & 252 & 252 \\ \hline %% p32 (4)
p36 \ (32) & 0.046 & 9.71   & 26.170  & 26.8 & 1645.7  & 384 & 11k  & 255 & 248 \\ \hline %% p32 (5)
p37 \ (46) & 0.010 & 3.48   & 27.970  & 38.4 & 709.6   & 275 & 4k   & 183 & 183 \\ \hline %% p46 (1)
p38 \ (46) & 0.022 & 63.59  & 31.420  & 19.1 & 1128.5  & 277 & 4k   & 184 & 184 \\ \hline %% p46 (3)
p39 \ (34) & 0.058 & 12.68  & 35.910  & 29.5 & 2140.1  & 408 & 15k  & 271 & 264 \\ \hline %% p34 (5)
p40 \ (34) & 0.040 & 15.41  & 36.030  & 31.7 & 2126.8  & 421 & 13k  & 268 & 268 \\ \hline %% p34 (4)
p41 \ (48) & 0.022 & 103.44 & 36.120  & 20.3 & 1359.2  & 289 & 5k   & 192 & 192 \\ \hline %% p48 (3)
p42 \ (36) & 0.054 & 17.65  & 37.100  & 31.5 & 2410.5  & 432 & 18k  & 287 & 280 \\ \hline %% p36 (5)
p43 \ (48) & 0.010 & 4.06   & 37.820  & 41.9 & 841.0   & 287 & 6k   & 191 & 191 \\ \hline %% p48 (1)
p44 \ (50) & 0.014 & 4.72   & 41.650  & 44.7 & 904.3   & 299 & 5k   & 199 & 199 \\ \hline %% p50 (1)
p45 \ (36) & 0.040 & 20.64  & 42.470  & 34.9 & 2397.2  & 446 & 18k  & 284 & 284 \\ \hline %% p36 (4)
p46 \ (50) & 0.022 & 131.65 & 45.770  & 21.6 & 1575.5  & 301 & 6k   & 200 & 207 \\ \hline %% p50 (3)
p47 \ (38) & 0.048 & 25.80  & 48.710  & 34.5 & 2553.2  & 456 & 21k  & 303 & 296 \\ \hline %% p38 (5)
p48 \ (40) & 0.034 & 60.81  & 55.480  & 40.0 & 2658.5  & 496 & 19k  & 316 & 316 \\ \hline %% p40 (4)
p49 \ (38) & 0.042 & 35.48  & 57.540  & 38.1 & 2586.7  & 471 & 28k  & 300 & 300 \\ \hline %% p38 (4)
p50 \ (42) & 0.053 & 47.82  & 72.320  & 40.6 & 2665.7  & 504 & 26k  & 335 & 328 \\ \hline %% p42 (5)
p51 \ (40) & 0.061 & 47.58  & 84.570  & 37.5 & 2727.0  & 480 & 35k  & 319 & 312 \\ \hline %% p40 (5)
p52 \ (42) & 0.039 & 53.22  & 101.080 & 43.7 & 2724.10 & 521 & 36k  & 332 & 332 \\ \hline %% p42 (4)
p53 \ (44) & 0.037 & 75.32  & 105.530 & 47.4 & 2817.1  & 546 & 36k  & 348 & 348 \\ \hline %% p44 (4)
p54 \ (44) & 0.049 & 62.14  & 114.780 & 42.6 & 2793.1  & 528 & 43k  & 351 & 344 \\ \hline %% p44 (5)
p55 \ (46) & 0.048 & 84.28  & --      & 46.0 & --      & 552 & --   & 367 & --  \\ \hline %% p46 (5)
p56 \ (46) & 0.034 & 95.31  & --      & 51.1 & --      & 571 & --   & 364 & --  \\ \hline %% p46 (4)
p57 \ (48) & 0.056 & 108.66 & --      & 49.5 & --      & 576 & --   & 383 & --  \\ \hline %% p48 (5)
p58 \ (48) & 0.038 & 120.24 & --      & 53.8 & --      & 596 & --   & 380 & --  \\ \hline %% p48 (4)
p59 \ (50) & 0.050 & 132.46 & --      & 51.6 & --      & 600 & --   & 399 & --  \\ \hline %% p50 (5)
p60 \ (50) & 0.040 & 147.93 & --      & 57.9 & --      & 621 & --   & 396 & --  \\ \hline %% p50 (4)
\end{tabular}}
\begin{flushleft} 
\footnotesize \ \ 
$^{\rm *}$
An average time of retrieving an activity schema (i.e., in this experiment among four 
learned activity schemata) for each task problem. The retrieval time increases linearly 
with the number of learned activity schemata for a specific task. 
\end{flushleft}
\label{tbl:stack_exp}
\vspace{-20pt}
\end{table}
%---------------------------------------------------------------------------
%---------------------------------------------------------------------------

%%%%%%%%%%%%%%%%%%%%%%%%%%%%%%%%%%%%%%%%%%%%%%%%%%%%%%%%%%
%%%%%%%%%%%%%%%%%%%%%%%%%%%%%%%%%%%%%%%%%%%%%%%%%%%%%%%%%%

%%%%%%%%%%%%%%%%%%%%%%%%%%%%%%%%%%%%%%%%%%%%%%%%%%%%%%%%%%
%%%%%%%%%%%%%%%%%%%%%%%%%%%%%%%%%%%%%%%%%%%%%%%%%%%%%%%%%%

%%%%%%%%%%%%%%%%%%%%%%%%%%%%%%%%%%%%%%%%%%%%%%%%%%%%%%%%%%%%%%%%%%%%%%%%%%%%%%%
\subsection{\textsc{ROVER}}\label{sec:experiments:rover}
\begin{figure}[!t]
  \centering
  \begin{subfigure}[b]{.58\linewidth}
  \centering
        \includegraphics[width=\linewidth]{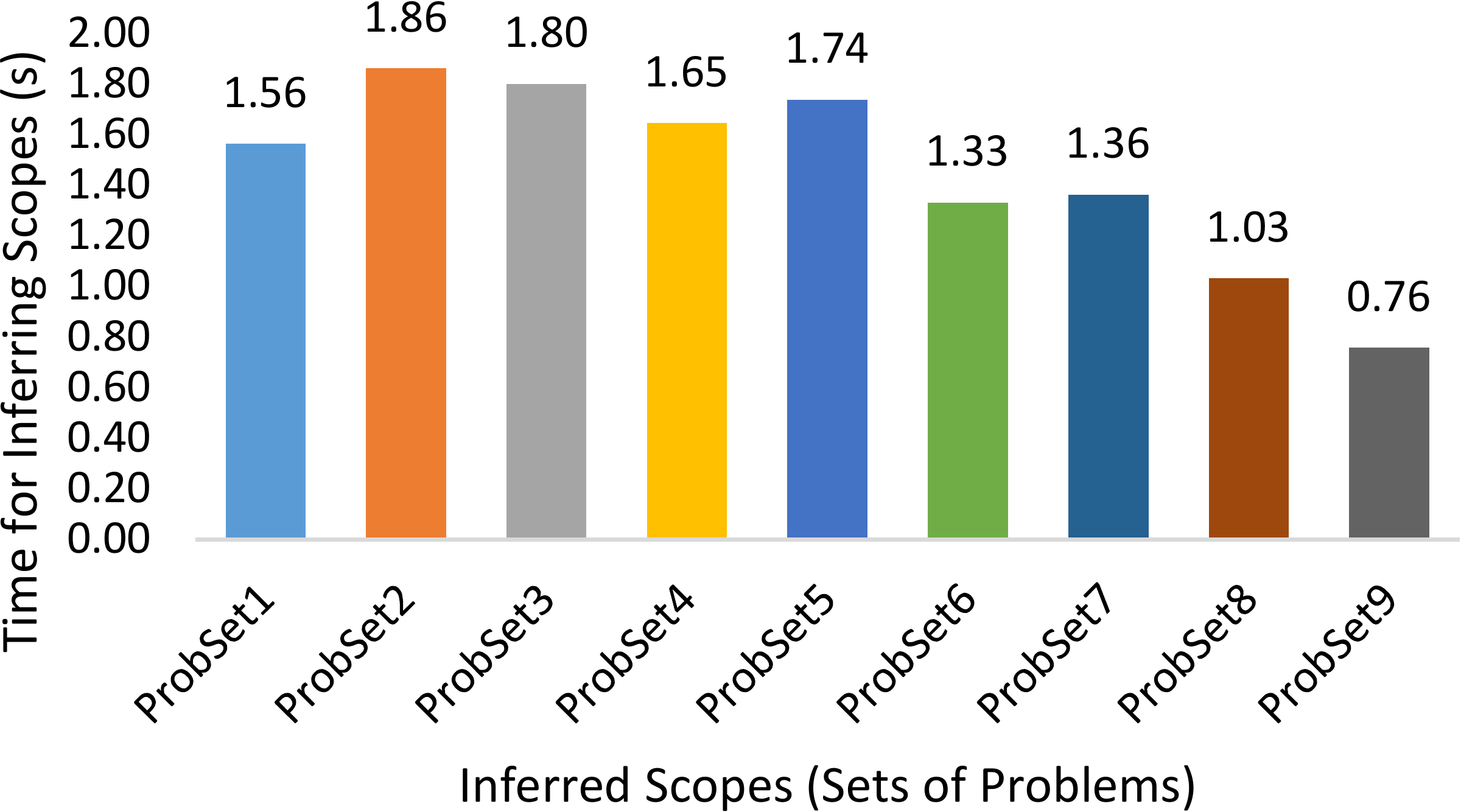}
%        \vspace{-15pt}
        \caption{} 
        \label{fig:rover_scope}
  \end{subfigure}\hspace{-2.4pt}
  \begin{subfigure}[b]{.41\linewidth}
  \centering
        \includegraphics[width=\linewidth]{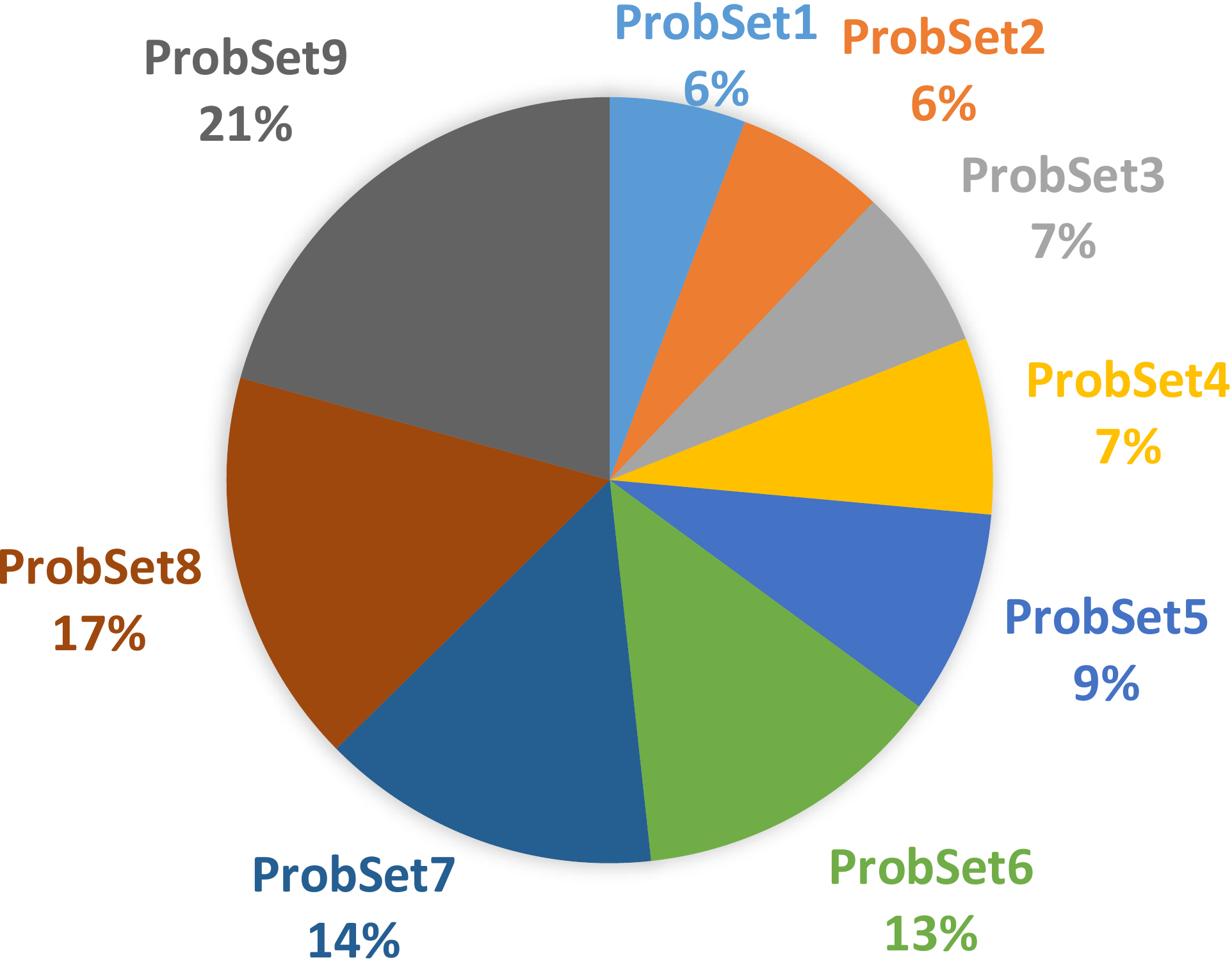}
%        \vspace{-15pt}
        \caption{} 
        \label{fig:rover_portion}
  \end{subfigure}\\
  %\vspace{-10pt}
  \caption{CPU time used by TVLA to classify the problems (a), 
  and distribution of the problems in the obtained problem sets 
  (b) in the \Rover domain}
  \label{fig:rover}
 \vspace{-5pt}
\end{figure}

\begin{figure}[!t]
  \centering
  \includegraphics[width=\linewidth]{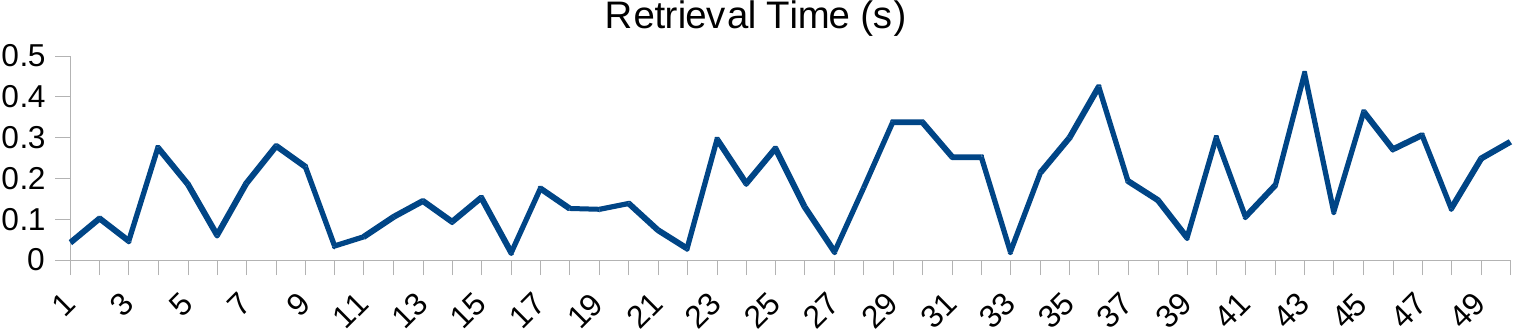}
  \caption{CPU time used by SBP to find an applicable activity 
  schema (among 9) for solving problems in the \Rover domain.}
  \label{fig:rover_retrieval}
 \vspace{-7pt}
\end{figure}

In the second experiment, we used the \Rover domain from the 
3rd International Planning Competition (IPC-3). 
In this experiment, we adopt a different approach for evaluating 
the proposed scope inference technique. 
We randomly generated $50$ problems containing exactly $1$ rover 
and ranging from $1$ to $3$ waypoints, $5$ to $30$ objectives, $5$ 
to $10$ cameras and $5$ to $20$ goals in each problem. 
Using the scope inference procedure, the problems are classified 
into $9$ sets of problems. That is, problems that converge to the 
same $3$-valued structure are put together in the same set. 
Hence, each set of problems is identified with a distinct scope 
of applicability. 
Figure~\ref{fig:rover_scope} shows the time required to classify 
the problems into different sets, i.e., the time required by TVLA 
to generate $3$-valued structures for the problems and test 
which problems converge to the same $3$-valued structure. 
Figure~\ref{fig:rover_portion} shows the distribution of the 
problems in the obtained sets of problems. 
In each set of problems, we simulated an experience and generated 
an activity schema for problem solving. 
Figure~\ref{fig:rover_retrieval} shows the time required to retrieve 
an applicable activity schema (among 9 activity schemata in this 
experiment) for solving given problems, i.e., the time required to 
check whether a given problem is embedded in the scope of an activity 
schema. 
% Using the obtained activity schemata, 
SBP successfully solved all problems in each class. 

%%%%%%%%%%%%%%%%%%%%%%%%%%%%%%%%%%%%%%%%%%%%%%%%%%%%%%%%%%
%%%%%%%%%%%%%%%%%%%%%%%%%%%%%%%%%%%%%%%%%%%%%%%%%%%%%%%%%%

%%%%%%%%%%%%%%%%%%%%%%%%%%%%%%%%%%%%%%%%%%%%%%%%%%%%%%%%%%
%%%%%%%%%%%%%%%%%%%%%%%%%%%%%%%%%%%%%%%%%%%%%%%%%%%%%%%%%%

%---------------------------------------------------------------------------
\subsection{\textsc{CAFE}}\label{sec:experiments:real}

\begin{figure*}[!t]
    \centering
    \begin{subfigure}[b]{.49\textwidth}
        \includegraphics[width=\textwidth]{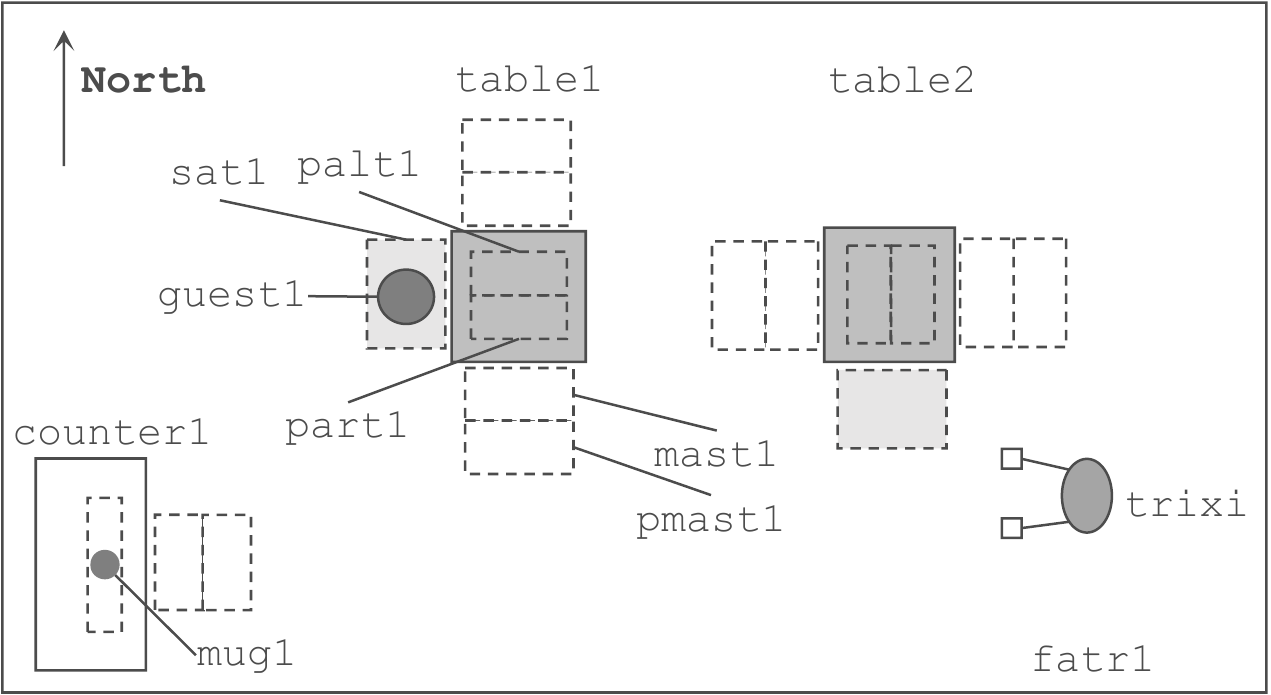}
        \caption{Scenario A}
        % \label{fig:concrete_structure}
    \end{subfigure}
    \ \ 
    \begin{subfigure}[b]{.49\textwidth}
        \includegraphics[width=\textwidth]{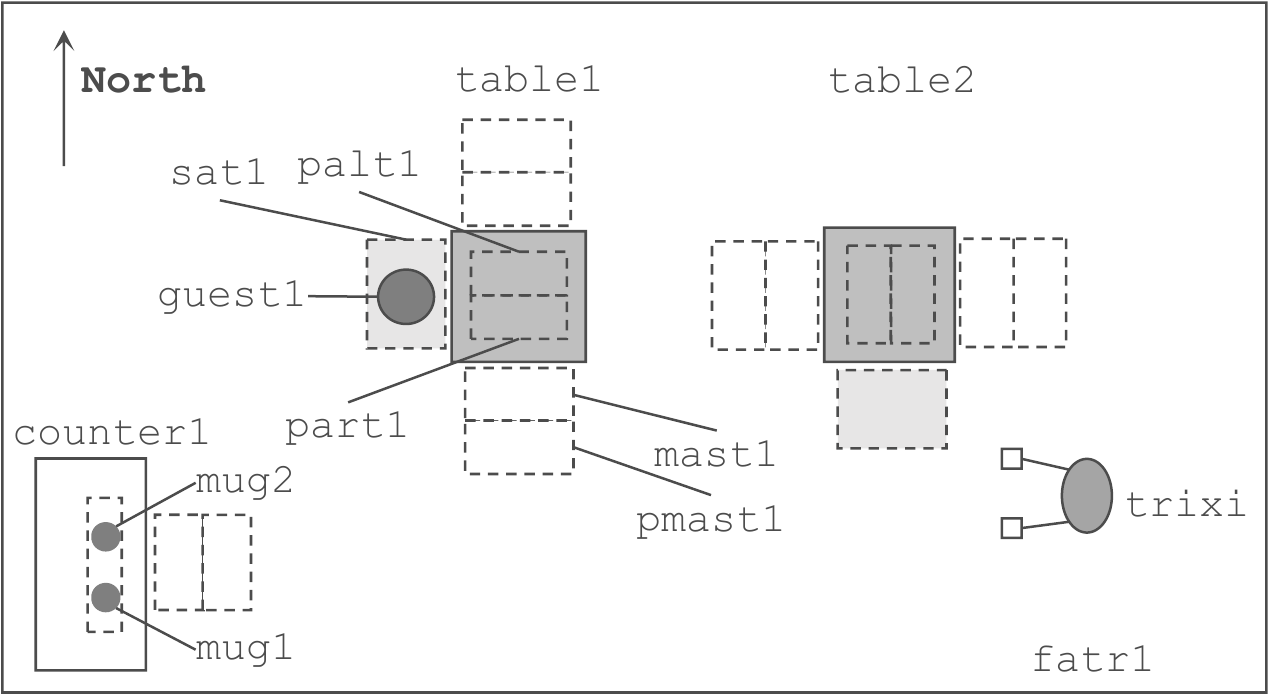}
        \caption{Scenario B}
        % \label{fig:abstract_structure}
    \end{subfigure}
    \caption{Initial states of the restaurant floor for the \texttt{ServeACoffee} 
    demonstration in Scenarios A and B with a PR2. (a) In Scenario A, 
    PR2 is taught to take \texttt{mug1} from \texttt{counter1} and approaches the 
    south of \texttt{table1} and place \texttt{mug1} on the right side of \texttt{guest1}.
    (b) In Scenario B, PR2 is taught to take \texttt{mug1} and \texttt{mug2} 
    from \texttt{counter1} and approaches the south of \texttt{table1} and place 
    \texttt{mug1} on the right side of \texttt{guest1}, and then approaches the north 
    of \texttt{table1} and place \texttt{mug2} on the left side of \texttt{guest1}.}
    \label{fig:cafe_scenarios}
\end{figure*}

\begin{figure*}[!t]
    \centering
    \begin{subfigure}[b]{.246\textwidth}
        \includegraphics[width=\textwidth]{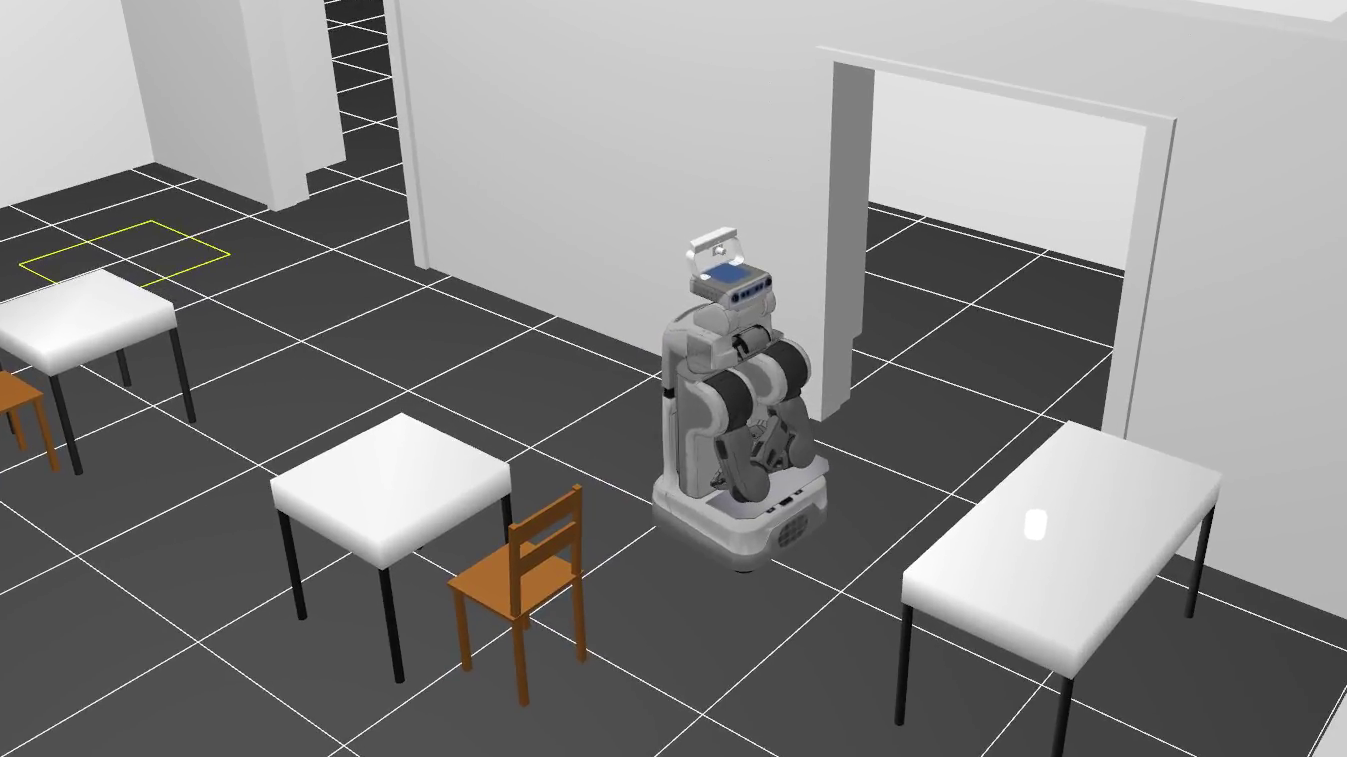}
    \end{subfigure}\hspace{-1pt}
    \begin{subfigure}[b]{.246\textwidth}
        \includegraphics[width=\textwidth]{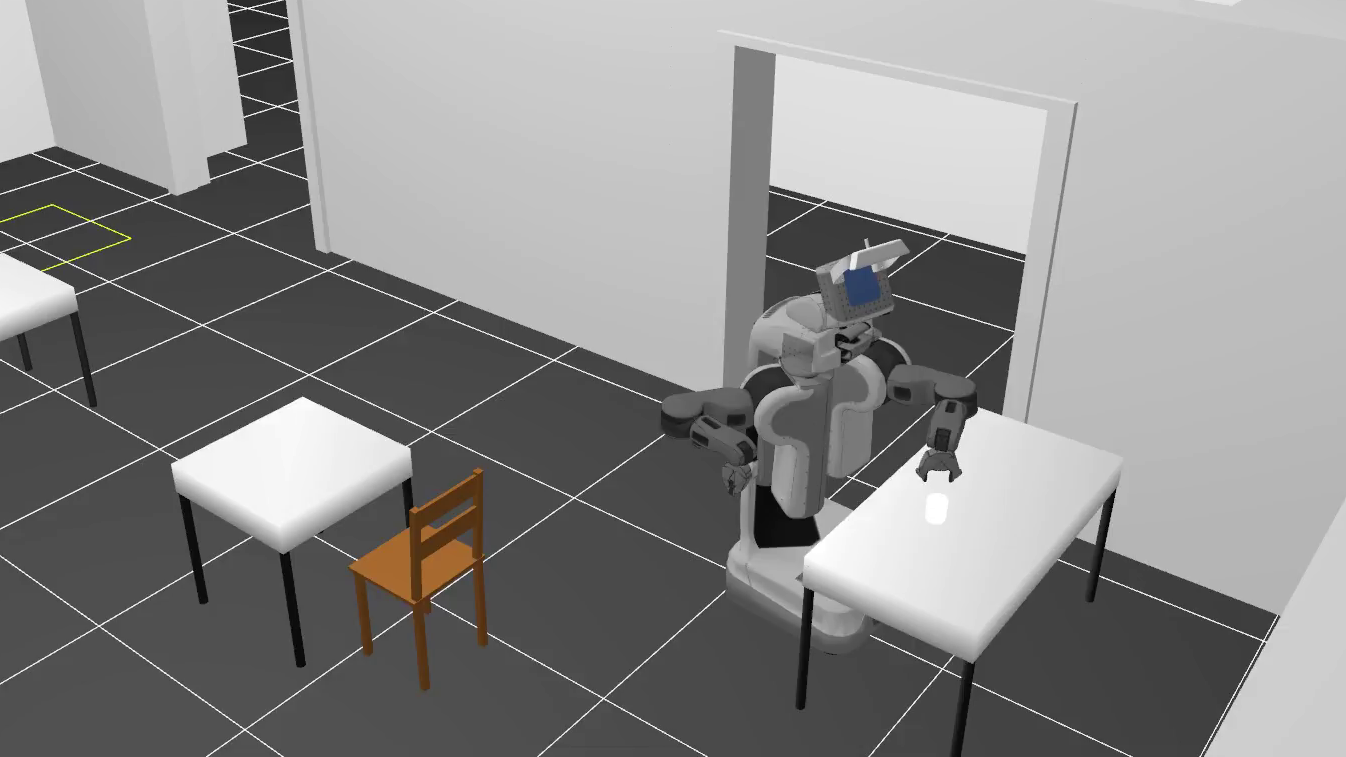}
    \end{subfigure}\hspace{-1pt}
    \begin{subfigure}[b]{.246\textwidth}
        \includegraphics[width=\textwidth]{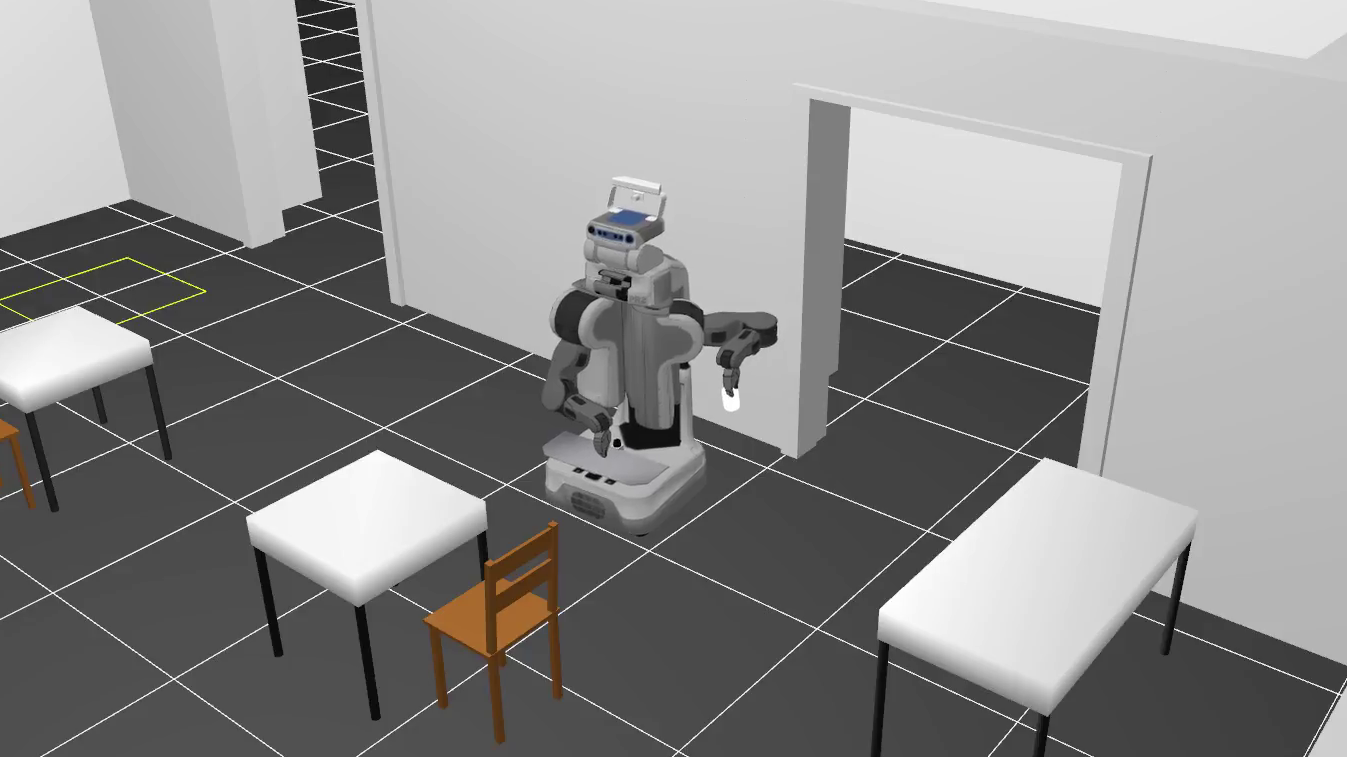}
    \end{subfigure}\hspace{-1pt}
    \begin{subfigure}[b]{.246\textwidth}
        \includegraphics[width=\textwidth]{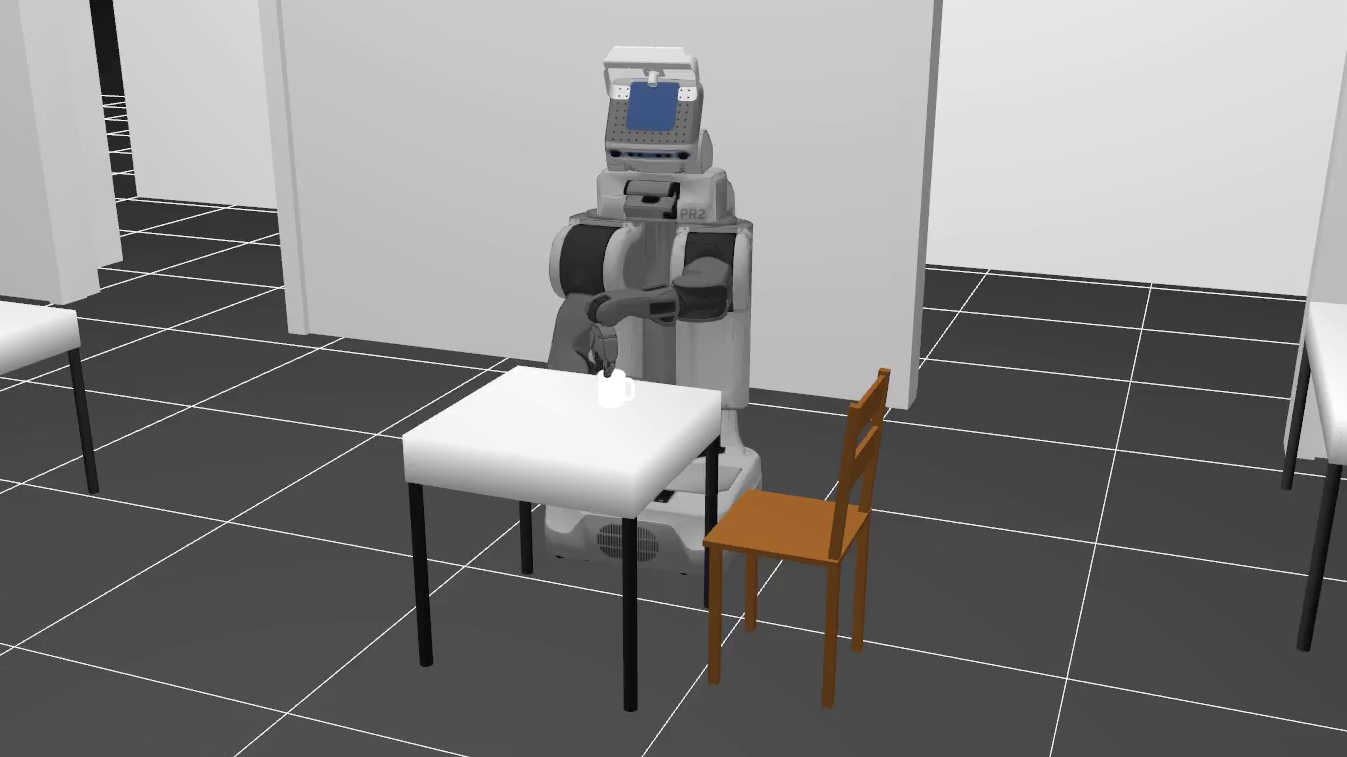}
    \end{subfigure}\hspace{-1pt}
    \caption{An example of the execution of a \texttt{ServeACoffee} task with a 
    PR2 in Gazebo simulated environment. 
    In this scenario, (from left to right) robot moves to a \textit{counter}, 
    picks up a \textit{mug} from the counter, moves to a \textit{table}, and 
    puts the mug on the table in front of a \textit{guest}.}
    \label{fig:sim_cafe}
\end{figure*}

In order to validate the practical utility of our approach, we also applied 
it to a real-world task using a fully physically simulated PR2 in Gazebo, 
the standard simulator in ROS. 
We developed a \Cafe EBPD including $14$ concrete and 
$4$ abstract planning operators (see Table~\ref{tbl:cafe_domain}).
We use a coffee serving demonstration including two Scenarios A and B with 
different sets of instructions to teach a PR2 to serve a guest in a 
cafe environment (see Figure~\ref{fig:cafe_scenarios}). 
Instructions for Scenario A is ``Move to counter1, grasp mug1, move to south 
of table1, place mug1 at the right placement area of guest1 -- this is a 
\texttt{ServeACoffee} task.''
Instructions for Scenario B is ``Move to counter1, grasp mug1 and mug2, 
move to south of table1, place mug1 at the right placement area of guest1, 
move to north of table1, place mug2 at the left placement area of guest1 
-- this is also a \texttt{ServeACoffee} task.''
In both scenarios, it is assumed that the robot knows the location of 
the guest and of the placement areas on the table. However, it does not 
know which placement area to approach for a guest. 
We used the infrastructure and simulation environment of the RACE 
project \footnote{ \texttt{{http://project-race.eu/}}} \cite{race2014} 
for instruction-based teaching of the robot to achieve the tasks. 
Figure~\ref{fig:sim_cafe} shows the snapshots of teaching the PR2 
a \texttt{ServeACoffee} task in Scenario A in Gazebo. 

%%%%%%%%%%%%%%%%%%%%%%%%%%%%%
\begin{table}[t]
\centering
\caption{Abstract and planning operators in the \Cafe EBPD. 
}
% \resizebox{\columnwidth}{!}
{\footnotesize
\begin{tabular}[l]{rl}
% \hline
\textbf{\normalsize Abstract operators} & \textbf{\normalsize Concrete operators}\\
\hline
\texttt{move/3}      & \texttt{move-base/3} \\
\texttt{move/3}      & \texttt{move-base-blind/3} \\
\texttt{pick/4}      & \texttt{pick-up-object/8} \\
\texttt{place/4}     & \texttt{place-object/8} \\
$\varnothing$ & \texttt{tuck-arm/5} \\
$\varnothing$ & \texttt{move-arm-to-carry/5} \\
$\varnothing$ & \texttt{move-arm-to-side/5} \\
$\varnothing$ & \texttt{move-torso-down/5} \\
$\varnothing$ & \texttt{move-torso-middle/5} \\
$\varnothing$ & \texttt{move-torso-up/5} \\
$\varnothing$ & \texttt{ready-to-safe-move-with-no-object/8} \\
$\varnothing$ & \texttt{ready-to-safe-move-with-one-object/9} \\
$\varnothing$ & \texttt{ready-to-safe-move-with-two-object/10} \\
$\varnothing$ & \texttt{observe-object-on-area/4} \\
% \hline
\end{tabular}}
\label{tbl:cafe_domain}
\end{table}
%%%%%%%%%%%%%%%%%%%%%%%%%%%%%

%%%%%%%%%%%%%%%%%%%%%%%%%%%%%%%%%%%%%%%%%%%%%%%%%%%%%%%%%%
%%%%%%%%%%%%%%%%%%%%%%%%%%%%%%%%%%%%%%%%%%%%%%%%%%%%%%%%%%

%%%%%%%%%%%%%%%%%%%%%%%%%%%%%%%%%%%%%%%%%%%%%%%%%%%%%%%%%%
%%%%%%%%%%%%%%%%%%%%%%%%%%%%%%%%%%%%%%%%%%%%%%%%%%%%%%%%%%

Our system learned two activity schemata for \texttt{ServeACoffee} task with 
distinct abstract plans (i.e, different instructions sets) and distinct 
scopes of applicability. 
To validate the utility of the learned activity schemata, 
we setup two test scenarios in each class of \texttt{ServeACoffee} task 
in which the robot is asked for serving a guest sitting at \texttt{table2}.
Our system computes the solution plans for each task problem using the 
learned activity schemata in less than $1$ second. 
Video of the PR2 doing \texttt{ServeACoffee} tasks in this experiment 
are available at \url{https://goo.gl/HJ6g2R} and 
\url{https://github.com/mokhtarivahid/ebpd/tree/master/demos}.

The domains, experiences, learned activity schemata and given task 
problems used in our experiments are available online by the link: 
\url{https://github.com/mokhtarivahid/ebpd/}.

%%%%%%%%%%%%%%%%%%%%%%%%%%%%%%%%%%%%%%%%%%%%%%%%%%%%%%%%%%%%%%%%%%%%%%%%%%
\section{CONCLUSION AND FUTURE WORK}\label{sec:conclusion}

We proposed an approach to generate a set of conditions that determines the 
scope of applicability of an activity schema in experience-based planning 
domains (EBPDs). 
The inferred scope allows an EBPD system to automatically find an applicable 
activity schema for solving a task problem, among several learned activity 
schemata for a specific task. 

We validated the utility of this work in a simulated domain and a fully 
physically simulated PR2 in Gazebo. Through our experiments, we demonstrated 
the effectiveness of the system, including loop detection and scope inference 
procedures. We showed the timing results for test problems in these experiments. 
The time required for learning activity schemata, and computing and testing 
their scopes were negligible. 
The system learned activity schemata from single examples in under seconds, 
in contrast to other machine learning techniques, addressed in the related 
work, which usually require large sets of plan traces to learn planning 
domain knowledge (e.g., HTN-Maker \citep{hogg2008htn} uses 75 out of 100 
input problems to train the system). 

While the results show good scalability, many engineering optimizations are 
possible on our prototype implementation of the proposed algorithms. Faster 
results can be obtained from an implementation in a compiled language. Extensive 
evaluation of the proposed system on a large set of domains is also part of 
the future work.

%%%%%%%%%%%%%%%%%%%%%%%%%%%%%%%%%%%%%%%%%%%%%%%%%%%%%%%%%%%
%%%%%%%%%%%%%%%%%%%%%%%%%%%%%%%%%%%%%%%%%%%%%%%%%%%%%%%%%%%
%%%%%%%%%%%%%%%%%%%%%%%%%%%%%%%%%%%%%%%%%%%%%%%%%%%%%%%%%%%
%%%%%%%%%%%%%%%%%%%%%%%%%%%%%%%%%%%%%%%%%%%%%%%%%%%%%%%%%%%
%%%%%%%%%%%%%%%%%%%%%%%%%%%%%%%%%%%%%%%%%%%%%%%%%%%%%%%%%%%
%%%%%%%%%%%%%%%%%%%%%%%%%%%%%%%%%%%%%%%%%%%%%%%%%%%%%%%%%%%
%%%%%%%%%%%%%%%%%%%%%%%%%%%%%%%%%%%%%%%%%%%%%%%%%%%%%%%%%%%

\begin{acks}
%%%%%%%%%%%%%%%%%%%%%%%%%%%%%%%%%%%%%%%%%%%%%%%%%%%%%%%%%%%%%%%%%%%%%%%%%%%%%%%%
This work is funded by  the Portuguese Foundation for Science 
and Technology (FCT) under the grant SFRH/BD/94184/2013,
and the project UID/CEC/00127/\-2013.
% We are also thankful to Prof. Daniele Magazzeni, for hosting 
% this work at King's College London.

\end{acks}

% Bibliography
\bibliographystyle{ACM-Reference-Format}
\bibliography{ref}


\begin{thebibliography}{30}

%%% the Web address.

\ifx \showCODEN    \undefined \def \showCODEN     #1{\unskip}     \fi
\ifx \showDOI      \undefined \def \showDOI       #1{#1}\fi
\ifx \showISBNx    \undefined \def \showISBNx     #1{\unskip}     \fi
\ifx \showISBNxiii \undefined \def \showISBNxiii  #1{\unskip}     \fi
\ifx \showISSN     \undefined \def \showISSN      #1{\unskip}     \fi
\ifx \showLCCN     \undefined \def \showLCCN      #1{\unskip}     \fi
\ifx \shownote     \undefined \def \shownote      #1{#1}          \fi
\ifx \showarticletitle \undefined \def \showarticletitle #1{#1}   \fi
\ifx \showURL      \undefined \def \showURL       {\relax}        \fi
% invisible to TeX
\providecommand\bibfield[2]{#2}
\providecommand\bibinfo[2]{#2}
\providecommand\natexlab[1]{#1}
\providecommand\showeprint[2][]{arXiv:#2}

\bibitem[\protect\citeauthoryear{Allen, Chambers, Ferguson, Galescu, Jung,
  Swift, and Taysom}{Allen et~al\mbox{.}}{2007}]%
        {allen2007plow}
\bibfield{author}{\bibinfo{person}{James Allen}, \bibinfo{person}{Nathanael
  Chambers}, \bibinfo{person}{George Ferguson}, \bibinfo{person}{Lucian
  Galescu}, \bibinfo{person}{Hyuckchul Jung}, \bibinfo{person}{Mary Swift},
  {and} \bibinfo{person}{William Taysom}.} \bibinfo{year}{2007}\natexlab{}.
\newblock \showarticletitle{{PLOW}: A collaborative task learning agent}. In
  \bibinfo{booktitle}{\emph{AAAI}}, Vol.~\bibinfo{volume}{7}.
  \bibinfo{pages}{1514--1519}.
\newblock


\bibitem[\protect\citeauthoryear{Argall, Chernova, Veloso, and Browning}{Argall
  et~al\mbox{.}}{2009}]%
        {Argall2009469}
\bibfield{author}{\bibinfo{person}{Brenna~D. Argall}, \bibinfo{person}{Sonia
  Chernova}, \bibinfo{person}{Manuela Veloso}, {and} \bibinfo{person}{Brett
  Browning}.} \bibinfo{year}{2009}\natexlab{}.
\newblock \showarticletitle{A survey of robot learning from demonstration}.
\newblock \bibinfo{journal}{\emph{Robotics and Autonomous Systems}}
  \bibinfo{volume}{57}, \bibinfo{number}{5} (\bibinfo{year}{2009}),
  \bibinfo{pages}{469 -- 483}.
\newblock
\showISSN{0921-8890}


\bibitem[\protect\citeauthoryear{Billard, Calinon, Dillmann, and
  Schaal}{Billard et~al\mbox{.}}{2008}]%
        {billard2008robot}
\bibfield{author}{\bibinfo{person}{Aude Billard}, \bibinfo{person}{Sylvain
  Calinon}, \bibinfo{person}{Ruediger Dillmann}, {and} \bibinfo{person}{Stefan
  Schaal}.} \bibinfo{year}{2008}\natexlab{}.
\newblock \showarticletitle{Robot programming by demonstration}.
\newblock In \bibinfo{booktitle}{\emph{Springer handbook of robotics}}.
  \bibinfo{publisher}{Springer}, \bibinfo{pages}{1371--1394}.
\newblock


\bibitem[\protect\citeauthoryear{Borrajo, Roub\'{\i}\v{c}kov\'{a}, and
  Serina}{Borrajo et~al\mbox{.}}{2015}]%
        {borrajo2015acm}
\bibfield{author}{\bibinfo{person}{Daniel Borrajo}, \bibinfo{person}{Anna
  Roub\'{\i}\v{c}kov\'{a}}, {and} \bibinfo{person}{Ivan Serina}.}
  \bibinfo{year}{2015}\natexlab{}.
\newblock \showarticletitle{Progress in case-based planning}.
\newblock \bibinfo{journal}{\emph{ACM Computing Surveys (CSUR)}}
  \bibinfo{volume}{47}, \bibinfo{number}{2}, Article \bibinfo{articleno}{35}
  (\bibinfo{date}{Jan} \bibinfo{year}{2015}), \bibinfo{numpages}{39}~pages.
\newblock


\bibitem[\protect\citeauthoryear{Bylander}{Bylander}{1994}]%
        {bylander1994computational}
\bibfield{author}{\bibinfo{person}{Tom Bylander}.}
  \bibinfo{year}{1994}\natexlab{}.
\newblock \showarticletitle{The computational complexity of propositional
  {STRIPS} planning}.
\newblock \bibinfo{journal}{\emph{Artificial Intelligence}}
  \bibinfo{volume}{69}, \bibinfo{number}{1-2} (\bibinfo{year}{1994}),
  \bibinfo{pages}{165--204}.
\newblock


\bibitem[\protect\citeauthoryear{Chao, Cakmak, and Thomaz}{Chao
  et~al\mbox{.}}{2011}]%
        {chao2011towards}
\bibfield{author}{\bibinfo{person}{Crystal Chao}, \bibinfo{person}{Maya
  Cakmak}, {and} \bibinfo{person}{Andrea~L Thomaz}.}
  \bibinfo{year}{2011}\natexlab{}.
\newblock \showarticletitle{Towards grounding concepts for transfer in goal
  learning from demonstration}. In \bibinfo{booktitle}{\emph{Development and
  Learning (ICDL), 2011 IEEE International Conference on}},
  Vol.~\bibinfo{volume}{2}. IEEE, \bibinfo{pages}{1--6}.
\newblock


\bibitem[\protect\citeauthoryear{Chrpa}{Chrpa}{2010}]%
        {chrpa2010generation}
\bibfield{author}{\bibinfo{person}{Luk{\'a}{\v{s}} Chrpa}.}
  \bibinfo{year}{2010}\natexlab{}.
\newblock \showarticletitle{Generation of macro-operators via investigation of
  action dependencies in plans}.
\newblock \bibinfo{journal}{\emph{The Knowledge Engineering Review}}
  \bibinfo{volume}{25}, \bibinfo{number}{03} (\bibinfo{year}{2010}),
  \bibinfo{pages}{281--297}.
\newblock


\bibitem[\protect\citeauthoryear{Fawcett and Utgoff}{Fawcett and
  Utgoff}{1992}]%
        {fawcett1992automatic}
\bibfield{author}{\bibinfo{person}{Tom~E Fawcett} {and} \bibinfo{person}{Paul~E
  Utgoff}.} \bibinfo{year}{1992}\natexlab{}.
\newblock \showarticletitle{Automatic feature generation for problem solving
  systems}.
\newblock In \bibinfo{booktitle}{\emph{Machine Learning Proceedings 1992}}.
  \bibinfo{publisher}{Elsevier}, \bibinfo{pages}{144--153}.
\newblock


\bibitem[\protect\citeauthoryear{Fikes, Hart, and Nilsson}{Fikes
  et~al\mbox{.}}{1972}]%
        {fikes1972strips2}
\bibfield{author}{\bibinfo{person}{Richard~E Fikes}, \bibinfo{person}{Peter~E.
  Hart}, {and} \bibinfo{person}{Nils~J Nilsson}.}
  \bibinfo{year}{1972}\natexlab{}.
\newblock \showarticletitle{Learning and executing generalized robot plans}.
\newblock \bibinfo{journal}{\emph{Artificial intelligence}}
  \bibinfo{volume}{3} (\bibinfo{year}{1972}), \bibinfo{pages}{251--288}.
\newblock


\bibitem[\protect\citeauthoryear{Ghallab, Nau, and Traverso}{Ghallab
  et~al\mbox{.}}{2004}]%
        {ghallab2004automated}
\bibfield{author}{\bibinfo{person}{Malik Ghallab}, \bibinfo{person}{Dana Nau},
  {and} \bibinfo{person}{Paolo Traverso}.} \bibinfo{year}{2004}\natexlab{}.
\newblock \bibinfo{booktitle}{\emph{Automated planning: theory \& practice}}.
\newblock \bibinfo{publisher}{Elsevier}.
\newblock


\bibitem[\protect\citeauthoryear{Hammond}{Hammond}{1986}]%
        {hammond1986chef}
\bibfield{author}{\bibinfo{person}{Kristian~J Hammond}.}
  \bibinfo{year}{1986}\natexlab{}.
\newblock \showarticletitle{{CHEF}: a model of case-based planning}. In
  \bibinfo{booktitle}{\emph{Proceedings of the Fifth National Conference on
  Artificial Intelligence}}. \bibinfo{publisher}{AAAI Press},
  \bibinfo{pages}{267--271}.
\newblock


\bibitem[\protect\citeauthoryear{Hertzberg, Zhang, Zhang, Rockel, Neumann,
  Lehmann, Dubba, Cohn, Saffiotti, Pecora, Mansouri, Kone\u{c}n\'{y},
  G{\"u}nther, Stock, {Seabra Lopes}, Oliveira, Lim, Kasaei, Mokhtari, Hotz,
  and Bohlken}{Hertzberg et~al\mbox{.}}{2014}]%
        {race2014}
\bibfield{author}{\bibinfo{person}{Joachim Hertzberg}, \bibinfo{person}{Jianwei
  Zhang}, \bibinfo{person}{Liwei Zhang}, \bibinfo{person}{Sebastian Rockel},
  \bibinfo{person}{Bernd Neumann}, \bibinfo{person}{Jos Lehmann},
  \bibinfo{person}{KrishnaS.R. Dubba}, \bibinfo{person}{AnthonyG. Cohn},
  \bibinfo{person}{Alessandro Saffiotti}, \bibinfo{person}{Federico Pecora},
  \bibinfo{person}{Masoumeh Mansouri}, \bibinfo{person}{{\v S}tefan
  Kone\u{c}n\'{y}}, \bibinfo{person}{Martin G{\"u}nther},
  \bibinfo{person}{Sebastian Stock}, \bibinfo{person}{Lu\'{i}s {Seabra Lopes}},
  \bibinfo{person}{Miguel Oliveira}, \bibinfo{person}{GiHyun Lim},
  \bibinfo{person}{Hamidreza Kasaei}, \bibinfo{person}{Vahid Mokhtari},
  \bibinfo{person}{Lothar Hotz}, {and} \bibinfo{person}{Wilfried Bohlken}.}
  \bibinfo{year}{2014}\natexlab{}.
\newblock \showarticletitle{The {RACE} project}.
\newblock \bibinfo{journal}{\emph{KI - K{\"u}nstliche Intelligenz}}
  \bibinfo{volume}{28}, \bibinfo{number}{4} (\bibinfo{year}{2014}),
  \bibinfo{pages}{297--304}.
\newblock
\showISSN{1610-1987}


\bibitem[\protect\citeauthoryear{Hogg, Munoz-Avila, and Kuter}{Hogg
  et~al\mbox{.}}{2008}]%
        {hogg2008htn}
\bibfield{author}{\bibinfo{person}{Chad Hogg}, \bibinfo{person}{H{\'e}ctor
  Munoz-Avila}, {and} \bibinfo{person}{Ugur Kuter}.}
  \bibinfo{year}{2008}\natexlab{}.
\newblock \showarticletitle{{HTN-MAKER}: learning {HTN}s with minimal
  additional knowledge engineering required}. In
  \bibinfo{booktitle}{\emph{Proceedings of the Twenty-Third AAAI Conference on
  Artificial Intelligence}}. \bibinfo{publisher}{AAAI Press},
  \bibinfo{pages}{950--956}.
\newblock


\bibitem[\protect\citeauthoryear{Ilghami and Nau}{Ilghami and Nau}{2006}]%
        {ilghami2006hdl}
\bibfield{author}{\bibinfo{person}{Okhtay Ilghami} {and}
  \bibinfo{person}{Dana~S Nau}.} \bibinfo{year}{2006}\natexlab{}.
\newblock \showarticletitle{Learning to do {HTN} planning}. In
  \bibinfo{booktitle}{\emph{16st International Conference on Automated Planning
  and Scheduling (ICAPS)}}. \bibinfo{publisher}{AAAI Press},
  \bibinfo{pages}{390--393}.
\newblock


\bibitem[\protect\citeauthoryear{Ilghami, Nau, Munoz-Avila, and Aha}{Ilghami
  et~al\mbox{.}}{2002}]%
        {ilghami2002camel}
\bibfield{author}{\bibinfo{person}{Okhtay Ilghami}, \bibinfo{person}{Dana~S
  Nau}, \bibinfo{person}{H{\'e}ctor Munoz-Avila}, {and}
  \bibinfo{person}{David~W Aha}.} \bibinfo{year}{2002}\natexlab{}.
\newblock \showarticletitle{{CaMeL}: learning method preconditions for {HTN}
  planning}. In \bibinfo{booktitle}{\emph{Proceedings of the Sixth
  International Conference on Artificial Intelligence Planning Systems
  (AIPS)}}. \bibinfo{pages}{131--142}.
\newblock


\bibitem[\protect\citeauthoryear{Ilghami, Nau, Munoz-Avila, and Aha}{Ilghami
  et~al\mbox{.}}{2005}]%
        {ilghami2005learning}
\bibfield{author}{\bibinfo{person}{Okhtay Ilghami}, \bibinfo{person}{Dana~S
  Nau}, \bibinfo{person}{H{\'e}ctor Munoz-Avila}, {and}
  \bibinfo{person}{David~W Aha}.} \bibinfo{year}{2005}\natexlab{}.
\newblock \showarticletitle{Learning preconditions for planning from plan
  traces and {HTN} structure}.
\newblock \bibinfo{journal}{\emph{Computational Intelligence}}
  \bibinfo{volume}{21}, \bibinfo{number}{4} (\bibinfo{year}{2005}),
  \bibinfo{pages}{388--413}.
\newblock


\bibitem[\protect\citeauthoryear{Kleene}{Kleene}{1952}]%
        {kleene1952introduction}
\bibfield{author}{\bibinfo{person}{Stephen~Cole Kleene}.}
  \bibinfo{year}{1952}\natexlab{}.
\newblock \bibinfo{booktitle}{\emph{Introduction to metamathematics}}.
  Vol.~\bibinfo{volume}{483}.
\newblock \bibinfo{publisher}{D. Van Nostrand Co., Inc., New York, N. Y.}
\newblock


\bibitem[\protect\citeauthoryear{Lev-Ami and Sagiv}{Lev-Ami and Sagiv}{2000}]%
        {lev2000tvla}
\bibfield{author}{\bibinfo{person}{Tal Lev-Ami} {and} \bibinfo{person}{M
  Sagiv}.} \bibinfo{year}{2000}\natexlab{}.
\newblock \showarticletitle{TVLA: A framework for kleene logic based static
  analyses}.
\newblock \bibinfo{journal}{\emph{Master's thesis, Tel Aviv University}}
  (\bibinfo{year}{2000}).
\newblock


\bibitem[\protect\citeauthoryear{Lev{-}Ami and Sagiv}{Lev{-}Ami and
  Sagiv}{2000}]%
        {LAmiS:SAS00}
\bibfield{author}{\bibinfo{person}{Tal Lev{-}Ami} {and} \bibinfo{person}{Shmuel
  Sagiv}.} \bibinfo{year}{2000}\natexlab{}.
\newblock \showarticletitle{{TVLA:} {A} system for implementing static
  analyses}. In \bibinfo{booktitle}{\emph{Static Analysis, 7th International
  Symposium, {SAS} 2000, Santa Barbara, CA, USA, June 29 - July 1, 2000,
  Proceedings}}. \bibinfo{pages}{280--301}.
\newblock


\bibitem[\protect\citeauthoryear{Manber and Myers}{Manber and Myers}{1993}]%
        {manber1993suffix}
\bibfield{author}{\bibinfo{person}{Udi Manber} {and} \bibinfo{person}{Gene
  Myers}.} \bibinfo{year}{1993}\natexlab{}.
\newblock \showarticletitle{Suffix arrays: a new method for on-line string
  searches}.
\newblock \bibinfo{journal}{\emph{SIAM J. Comput.}} \bibinfo{volume}{22},
  \bibinfo{number}{5} (\bibinfo{year}{1993}), \bibinfo{pages}{935--948}.
\newblock


\bibitem[\protect\citeauthoryear{Mitchell, Keller, and Kedar-Cabelli}{Mitchell
  et~al\mbox{.}}{1986}]%
        {mitchell1986explanation}
\bibfield{author}{\bibinfo{person}{Tom~M. Mitchell},
  \bibinfo{person}{Richard~M. Keller}, {and} \bibinfo{person}{Smadar~T.
  Kedar-Cabelli}.} \bibinfo{year}{1986}\natexlab{}.
\newblock \showarticletitle{Explanation-based generalization: a unifying view}.
\newblock \bibinfo{journal}{\emph{Machine Learning}} \bibinfo{volume}{1},
  \bibinfo{number}{1} (\bibinfo{year}{1986}), \bibinfo{pages}{47--80}.
\newblock
\showISSN{0885-6125}


\bibitem[\protect\citeauthoryear{Mokhtari, Lim, {Seabra Lopes}, and
  Pinho}{Mokhtari et~al\mbox{.}}{2016a}]%
        {vahid2014experience}
\bibfield{author}{\bibinfo{person}{Vahid Mokhtari}, \bibinfo{person}{GiHyun
  Lim}, \bibinfo{person}{Lu\'{i}s {Seabra Lopes}}, {and}
  \bibinfo{person}{Armando~J. Pinho}.} \bibinfo{year}{2016}\natexlab{a}.
\newblock \showarticletitle{Gathering and conceptualizing plan-based robot
  activity experiences}.
\newblock In \bibinfo{booktitle}{\emph{Intelligent Autonomous Systems 13}},
  \bibfield{editor}{\bibinfo{person}{Emanuele Menegatti},
  \bibinfo{person}{Nathan Michael}, \bibinfo{person}{Karsten Berns}, {and}
  \bibinfo{person}{Hiroaki Yamaguchi}} (Eds.). \bibinfo{series}{Advances in
  Intelligent Systems and Computing}, Vol.~\bibinfo{volume}{302}.
  \bibinfo{publisher}{Springer International Publishing},
  \bibinfo{pages}{993--1005}.
\newblock


\bibitem[\protect\citeauthoryear{Mokhtari, {Seabra Lopes}, and Pinho}{Mokhtari
  et~al\mbox{.}}{2016b}]%
        {mokhtari2016jint}
\bibfield{author}{\bibinfo{person}{Vahid Mokhtari}, \bibinfo{person}{Lu\'{i}s
  {Seabra Lopes}}, {and} \bibinfo{person}{Armando~J. Pinho}.}
  \bibinfo{year}{2016}\natexlab{b}.
\newblock \showarticletitle{Experience-based planning domains: an integrated
  learning and deliberation approach for intelligent robots}.
\newblock \bibinfo{journal}{\emph{Journal of Intelligent {\&} Robotic Systems}}
  \bibinfo{volume}{83}, \bibinfo{number}{3} (\bibinfo{year}{2016}),
  \bibinfo{pages}{463--483}.
\newblock


\bibitem[\protect\citeauthoryear{Mokhtari, {Seabra Lopes}, and Pinho}{Mokhtari
  et~al\mbox{.}}{2016c}]%
        {mokhtari2016icaps}
\bibfield{author}{\bibinfo{person}{Vahid Mokhtari}, \bibinfo{person}{Lu\'{i}s
  {Seabra Lopes}}, {and} \bibinfo{person}{Armando~J. Pinho}.}
  \bibinfo{year}{2016}\natexlab{c}.
\newblock \showarticletitle{Experience-based robot task learning and planning
  with goal inference}. In \bibinfo{booktitle}{\emph{26st International
  Conference on Automated Planning and Scheduling (ICAPS)}}.
  \bibinfo{publisher}{AAAI Press}, \bibinfo{pages}{509--517}.
\newblock


\bibitem[\protect\citeauthoryear{Mokhtari, {Seabra Lopes}, and Pinho}{Mokhtari
  et~al\mbox{.}}{2017a}]%
        {vahid2017iros}
\bibfield{author}{\bibinfo{person}{Vahid Mokhtari}, \bibinfo{person}{Lu\'{i}s
  {Seabra Lopes}}, {and} \bibinfo{person}{Armando~J. Pinho}.}
  \bibinfo{year}{2017}\natexlab{a}.
\newblock \showarticletitle{An approach to robot task learning and planning
  with loops}. In \bibinfo{booktitle}{\emph{2017 IEEE/RSJ International
  Conference on Intelligent Robots and Systems (IROS)}}.
  \bibinfo{pages}{6033--6038}.
\newblock
\showISSN{2153-0866}


\bibitem[\protect\citeauthoryear{Mokhtari, {Seabra Lopes}, and Pinho}{Mokhtari
  et~al\mbox{.}}{2017b}]%
        {vahid2017prletter}
\bibfield{author}{\bibinfo{person}{Vahid Mokhtari}, \bibinfo{person}{Lu\'{i}s
  {Seabra Lopes}}, {and} \bibinfo{person}{Armando~J. Pinho}.}
  \bibinfo{year}{2017}\natexlab{b}.
\newblock \showarticletitle{Learning robot tasks with loops from experiences to
  enhance robot adaptability}.
\newblock \bibinfo{journal}{\emph{Pattern Recognition Letters}}
  \bibinfo{volume}{99}, \bibinfo{number}{Supplement C} (\bibinfo{year}{2017}),
  \bibinfo{pages}{57 -- 66}.
\newblock
\showISSN{0167-8655}
\urldef\tempurl%
\url{https://doi.org/10.1016/j.patrec.2017.06.003}
\showDOI{\tempurl}
\newblock
\shownote{User Profiling and Behavior Adaptation for Human-Robot Interaction.}


\bibitem[\protect\citeauthoryear{Rintanen}{Rintanen}{2012}]%
        {RINTANEN201245}
\bibfield{author}{\bibinfo{person}{Jussi Rintanen}.}
  \bibinfo{year}{2012}\natexlab{}.
\newblock \showarticletitle{Planning as satisfiability: heuristics}.
\newblock \bibinfo{journal}{\emph{Artificial Intelligence}}
  \bibinfo{volume}{193} (\bibinfo{year}{2012}), \bibinfo{pages}{45 -- 86}.
\newblock
\showISSN{0004-3702}


\bibitem[\protect\citeauthoryear{Sagiv, Reps, and Wilhelm}{Sagiv
  et~al\mbox{.}}{2002}]%
        {SRW:TOPLAS02}
\bibfield{author}{\bibinfo{person}{Shmuel Sagiv}, \bibinfo{person}{Thomas~W.
  Reps}, {and} \bibinfo{person}{Reinhard Wilhelm}.}
  \bibinfo{year}{2002}\natexlab{}.
\newblock \showarticletitle{Parametric shape analysis via 3-valued logic}.
\newblock \bibinfo{journal}{\emph{{ACM} Trans. Program. Lang. Syst.}}
  \bibinfo{volume}{24}, \bibinfo{number}{3} (\bibinfo{year}{2002}),
  \bibinfo{pages}{217--298}.
\newblock


\bibitem[\protect\citeauthoryear{Srivastava, Immerman, and
  Zilberstein}{Srivastava et~al\mbox{.}}{2011}]%
        {Srivastava2011615}
\bibfield{author}{\bibinfo{person}{Siddharth Srivastava}, \bibinfo{person}{Neil
  Immerman}, {and} \bibinfo{person}{Shlomo Zilberstein}.}
  \bibinfo{year}{2011}\natexlab{}.
\newblock \showarticletitle{A new representation and associated algorithms for
  generalized planning}.
\newblock \bibinfo{journal}{\emph{Artificial Intelligence}}
  \bibinfo{volume}{175}, \bibinfo{number}{2} (\bibinfo{year}{2011}),
  \bibinfo{pages}{615 -- 647}.
\newblock
\showISSN{0004-3702}


\bibitem[\protect\citeauthoryear{Winner and Veloso}{Winner and Veloso}{2007}]%
        {Winner07loopdistill}
\bibfield{author}{\bibinfo{person}{Elly Winner} {and}
  \bibinfo{person}{Manuela~M. Veloso}.} \bibinfo{year}{2007}\natexlab{}.
\newblock \showarticletitle{Loop{DISTILL}: learning domain-specific planners
  from example plans}. In \bibinfo{booktitle}{\emph{Workshop on {AI} Planning
  and Learning, {ICAPS}}}.
\newblock


\end{thebibliography}

\end{document}